\documentclass[lettersize,journal]{IEEEtran}
\usepackage{amsmath,amsfonts}
\usepackage{algorithmic}
\usepackage{algorithm}
\usepackage{array}
\usepackage[caption=false,font=normalsize,labelfont=sf,textfont=sf]{subfig}
\usepackage{textcomp}
\usepackage{stfloats}
\usepackage{url}
\usepackage{verbatim}
\usepackage{graphicx}
\usepackage{cite}
\usepackage{graphicx}
\usepackage[pdfborder={0 0 0}]{hyperref}
\usepackage{multirow}
\usepackage{makecell}
\begin{document}

\title{Geodesic Gradient Descent: A Generic and Learning-rate-free Optimizer on Objective Function-induced Manifolds}

\author{Liwei Hu\href{https://orcid.org/0000-0003-4994-9252}{\includegraphics[scale=0.5]{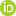}},
        Guangyao Li\href{https://orcid.org/0009-0009-1226-4007}{\includegraphics[scale=0.5]{ORCIDiD.png}},
        Wenyong Wang\href{https://orcid.org/0000-0003-4095-547X}{\includegraphics[scale=0.5]{ORCIDiD.png}},~\IEEEmembership{senior member,~IEEE,}
        Xiaoming Zhang\href{https://orcid.org/0000-0002-7091-1155}{\includegraphics[scale=0.5]{ORCIDiD.png}},
        Yu Xiang\href{https://orcid.org/0000-0001-9622-7661}{\includegraphics[scale=0.5]{ORCIDiD.png}},~\IEEEmembership{member,~IEEE}
\thanks{L. Hu, lecturer, is with the School of Information Science and Engineering, Hebei University of Science and Technology, Shijiazhuang 050091, China. (e-mail: \href{liweihu@hebust.edu.cn}{liweihu@hebust.edu.cn}).}
\thanks{G. Li, M.S. Candidate, is with the School of Information Science and Engineering, Hebei University of Science and Technology, Shijiazhuang 050091, China. (e-mail: \href{17743774386@163.com}{17743774386@163.com}).} 
\thanks{W. Wang, professor, is with the School of Computer Science and Engineering, University of Electronic Science and Technology of China, Chengdu 611731, China. (e-mail: \href{wangwy@uestc.edu.cn}{wangwy@uestc.edu.cn}).\\
W. Wang, special-term professor, is also with the International Institute of Next Generation Internet, Macau University of Science and Technology, Macau 519020, China.}%
\thanks{X. Zhang, professor, is with the School of Information Science and Engineering, Hebei University of Science and Technology, Shijiazhuang 050091, China. (e-mail: \href{zhangxiaom@hebust.edu.cn}{zhangxiaom@hebust.edu.cn}).} 
\thanks{Y. Xiang, professor, is with the School of Computer Science and Engineering, University of Electronic Science and Technology of China, Chengdu 611731, China. (e-mail: \href{jcxiang@uestc.edu.cn}{jcxiang@uestc.edu.cn}).}
\thanks{Corresponding authors: Xiaoming Zhang and Yu Xiang}
\thanks{This work is supported by the Youth Fund Project of the Department of Education of Hebei Province, China. No. QN2026857, and the Shijiazhuang Basic Research Program, No. 241790867A.}
\thanks{Manuscript received ***; revised ***.}}

\markboth{Journal of \LaTeX\ Class Files,~Vol.~14, No.~8, August~2021}%
{Shell \MakeLowercase{\textit{et al.}}: A Sample Article Using IEEEtran.cls for IEEE Journals}

\IEEEpubid{}

\maketitle

\begin{abstract}
Euclidean gradient descent algorithms barely capture the geometry of objective function-induced hypersurfaces and risk driving update trajectories off the hypersurfaces. Riemannian gradient descent algorithms address these issues but fail to represent complex hypersurfaces via a single classic manifold. We propose geodesic gradient descent (GGD), a generic and learning-rate-free Riemannian gradient descent algorithm. At each iteration, GGD uses an n-dimensional sphere to approximate a local neighborhood on the objective function-induced hypersurface, adapting to arbitrarily complex geometries. A tangent vector derived from the Euclidean gradient is projected onto the sphere to form a geodesic, ensuring the update trajectory stays on the hypersurface. Parameter updates are performed using the endpoint of the geodesic. The maximum step size of the gradient in GGD is equal to a quarter of the arc length on the n-dimensional sphere, thus eliminating the need for a learning rate. Experimental results show that compared with the classic Adam algorithm, GGD achieves test MSE reductions ranging from 35.79\% to 48.76\% for fully connected networks on the Burgers’ dataset, and cross-entropy loss reductions ranging from 3.14\% to 11.59\% for convolutional neural networks on the MNIST dataset.

\end{abstract}


\section{Introduction}
\begin{figure}
  \centering
  \includegraphics[scale=0.3]{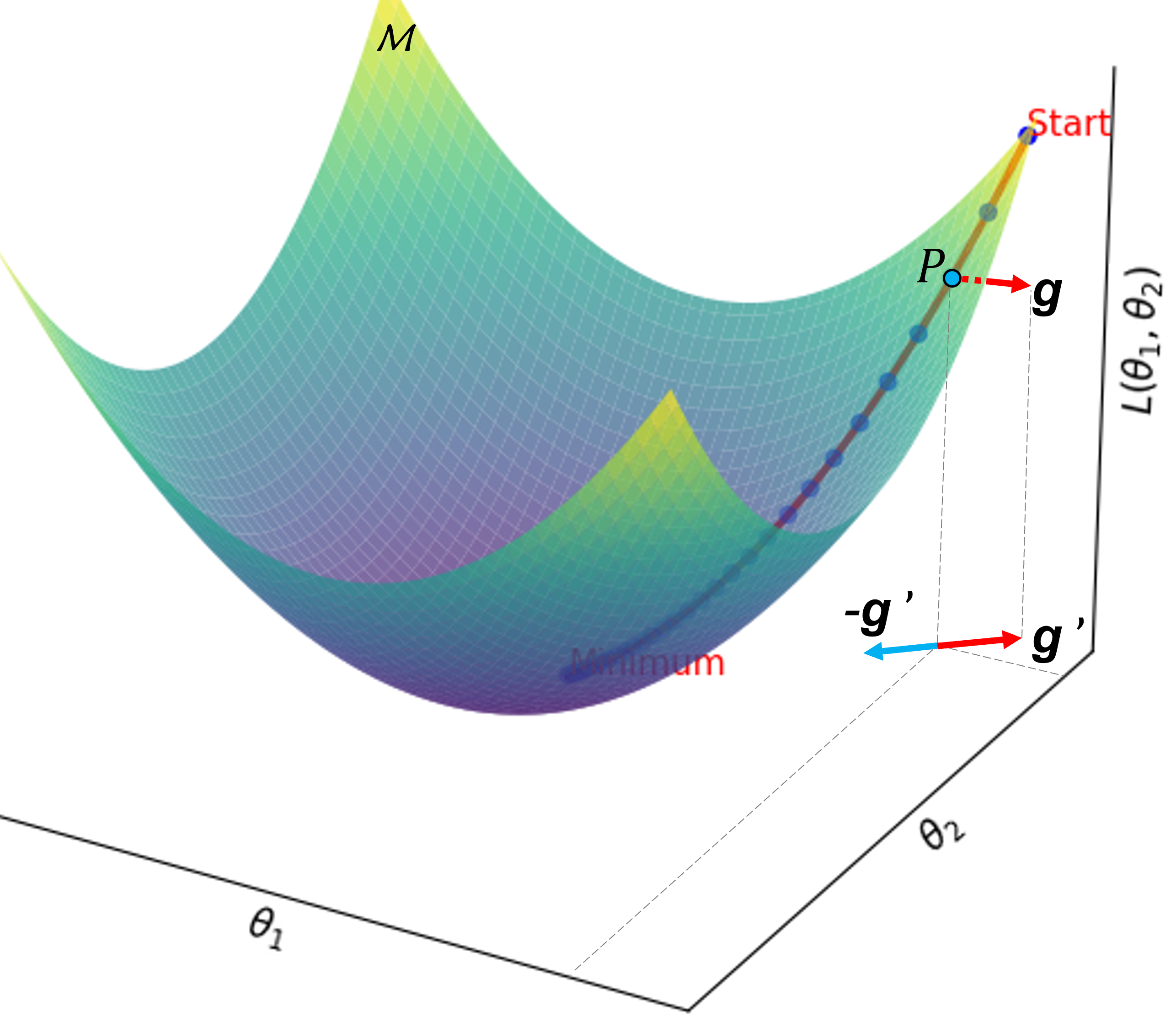}
  \caption{The convergence direction in Euclidean space.}
  \label{fig_gradients}
\end{figure}

\IEEEPARstart{G}{radient}-based algorithm is the most widely used optimization method in deep learning. Many practical problems in this field can be formulated as the optimization of an objective function with respect to (w.r.t) parameters:
\begin{equation}
\label{equ_optimization_E}
  \arg\min\limits_{\boldsymbol{\theta} \in \mathbb{R}^n} L(\boldsymbol{x};\boldsymbol{\theta})
\end{equation}
where $\boldsymbol{x}$ is the input data, $L(\boldsymbol{x};\boldsymbol{\theta})$ is the objective function determined by parameter $\boldsymbol{\theta}$. Classical gradient-based optimizers (e.g., stochastic gradient descent (SGD) \cite{deng2013recent}, adaptive moment (Adam) \cite{kingma2015adam}, adaptive subgradient (AdaGrad) \cite{duchi2011adaptive} and momentum orthogonalized by Newton-Schulz (Muon) \cite{jordan2024muon}) compute the gradient of the objective function w.r.t the parameters in Euclidean space to search for global optimal solutions. Figure \ref{fig_gradients} shows the convergence direction of gradient-based algorithms in Euclidean space. The gradient vector $\boldsymbol{g}=\{\frac{\partial L}{\partial \theta_1}, \frac{\partial L}{\partial \theta_2}, -1\}$ at point $P$ can be projected into the parameter space $\{\theta_1, \theta_2\}$ to obtain $\boldsymbol{g'}=\{\frac{\partial L}{\partial \theta_1}, \frac{\partial L}{\partial \theta_2}\}$. The negative direction of $\boldsymbol{g'}$ is the convergence direction at $P$.

The hypersurface (i.e., $\mathcal{M}$ in Figure \ref{fig_gradients}) induced by the objective function $L \colon \mathbb{R}^n\rightarrow \mathbb{R}$ is usually non-flat, whereas existing methods utilize Euclidean gradient vector to control the update trajectory of the point $P$ on $\mathcal{M}$. Since neither the convergence direction $-\boldsymbol{g'}$ nor the tangent vector at $P$ calculated according to $\boldsymbol{g}$ lies along the hypersurface $\mathcal{M}$, gradient-based optimizer in Euclidean space risks driving the update trajectory off the curved surface \cite{fei2025survey}. In addition, these optimizers in Euclidean space may ignore rich intrinsic geometric information (e.g., curvature and torsion) of the hypersurface.

Recent studies narrow down the search space from Euclidean space to a manifold \cite{boumal2023introduction, hu2020brief}, and thus Formula (\ref{equ_optimization_E}) can be reformulated as

\begin{equation}
\label{equ_optimization_M}
  \arg\min\limits_{\boldsymbol{\theta} \in \mathcal{M}} L(\boldsymbol{x};\boldsymbol{\theta})
\end{equation}
where $\mathcal{M}$, the hypersurface, is a manifold. In this case, performing optimization on $\mathcal{M}$ calculates the Riemannian gradient \cite{chen2026riemannian}, which enables the optimizers to regard the parameter space as a Riemannian manifold. The Riemannian gradient can be obtained by projecting the Euclidean gradient onto the tangent space of the manifold \cite{fei2025survey}, after which the Riemannian gradient can be mapped onto $M$ via a retraction function (e.g., exponential map) \cite{fei2025survey} to ensure that the update trajectory of point $P$ remains on $M$. Because the Riemannian gradient provides an accurate measurement of the objective function geometry (including the curvature, geodesics, torsion, and other intrinsic geometric properties) \cite{hu2025learning, fei2025survey,baker2024learning}, optimizers based on the Riemannian gradient are always suited for optimization tasks with manifold constrains. However, the objective function-induced hypersurface is usually a manifold with complex geometric structure, and it is difficult to constrain the parameter space using a single classic manifold. This is one of the reasons why the Riemannian gradient-based optimizer cannot achieve universality.

Motivated by Riemannian gradient, we propose a geodesic gradient descent (GGD) \footnote{The source code can be found at https://github.com/huliwei123/Geodesic-gradient-descent.git} algorithm that uses both the approximated Riemannian gradient and exponential map to perform gradient descent optimization on the objective function-induced hypersurface. In each iteration of the GGD algorithm, an n-dimensional (n-D) sphere that is tangent to the objective function-induced hypersurface at current parameter combination is used to approximate a small neighborhood of the hypersurface. This mechanism enables the GGD to adapt to arbitrarily complex geometry of objective functions. A tangent vector (i.e., an approximated Riemannian gradient) at the parameter combination is calculated from the Euclidean gradient. Then the tangent vector is projected onto the n-D sphere to form a geodesic with length equal to the norm of the tangent vector. Finally, the end point of the geodesic serves as the new parameter combination for the next iteration. In the GGD algorithm, the maximum update step size of parameters is equal to a quarter of the arc length on the sphere. Therefore, we eliminate the learning rate in this gradient-based optimization framework.

We highlight the contributions of this paper as follows: 1) We use an n-D sphere to approximate the objective function-induced hypersurface with complex geometric structure, thus proposing a generic geodesic gradient descent algorithm in Riemannian space; 2) In the GGD algorithm, we eliminate the learning rate, and instead, the maximum step size for parameter updates is determined by a quarter of the arc length on an n-D sphere; 3) The GGD algorithm achieves lower test errors in regression tasks and higher accuracy in classification tasks than exiting algorithms, e.g. Adam, SGD, and spherical SGD (SSGD) \cite{sadrtdinov2025sgd} etc.

The remainder of this paper is organized as follows. Section \ref{section_related_work} provides an overview of gradient-based optimizers in Euclidean space and Riemannian space. Section \ref{section_methodology} introduces the details of the proposed GGD algorithm. Section \ref{section_experimental_result} shows the experimental details and results of 6 different gradient-based algorithms on both regression and classification tasks. The conclusions are presented in Section \ref{section_conclusion}.

\section{RELATED WORKS}
\label{section_related_work}
In this section, we introduce the basic theory and key properties of gradient-based optimizers in Euclidean space and Riemannian space. 
\subsection{Gradient-based optimizers in Euclidean space}
The early classic gradient-based optimization algorithm is SGD \cite{deng2013recent}, which uses the gradient of the objective function w.r.t. the parameters for a single randomly selected data sample $\boldsymbol{x_i}$ to update the parameters:
\begin{equation}
\label{equ_SGD}
  \boldsymbol{\theta_{t+1}} = \boldsymbol{\theta_t}-\eta\nabla L(\boldsymbol{\theta_t};\boldsymbol{x_i}) 
\end{equation}
where $\eta$ denotes the learning rate. 

Since the gradient calculated from a single randomly selected data sample can not precisely reflect the overall gradient of the dataset, methods using a data subset to compute the average gradient for parameter updates (e.g., batch gradient descent (BGD) \cite{li2014efficient} and mini-batch gradient descent (MGD) \cite{khirirat2017mini}) typically achieve higher optimization accuracy:

\begin{equation}
\label{equ_SGD}
\begin{cases}
  \boldsymbol{g}_t = \frac{1}{N} \sum_{i=1}^{N} \nabla L(\boldsymbol{\theta_t};\boldsymbol{x_i})\\
  \boldsymbol{\theta_{t+1}} = \boldsymbol{\theta_t}-\eta \boldsymbol{g}_t
\end{cases}
\end{equation}
where $N$ denotes the number of samples in the subset. Although the aforementioned methods adopt the average gradient to correct the convergence direction, they still face the problem of being trapped in local optimal solutions. 

To alleviate the above problem of local optimal solutions, stochastic gradient descent with momentum (SGDM) was proposed \cite{ramezani2024generalization,sutskever2013importance}. SGDM modifies the average gradient of the current batch by continuously accumulating momentum, thereby enabling the model to escape from local optimal solutions. Although the convergence of SGDM is improved, the initial momentum is set to 0, leading to an obvious underestimated of the momentum. In addition, the SGDM algorithm fails to assign distinct learning rates to different parameters. Duchi et al. proposed the AdaGrad \cite{duchi2011adaptive} algorithm, which adaptively adjusts the learning rate for different parameters by the sum of squared historical gradients:
\begin{equation}
\label{equ_SGD}
\begin{cases}
    \boldsymbol{G}_t = \boldsymbol{G}_{t-1}+(\boldsymbol{g}_t)^2\\
    \boldsymbol{\theta}_{t+1} = \boldsymbol{\theta}_t - \frac{\eta}{\sqrt{\boldsymbol{G}_t+\epsilon}} \cdot \boldsymbol{g}_t.
\end{cases}
\end{equation}
Because $G_t$ is monotonically increasing, the learning rate decreases continuously (i.e., learning rate decay), which means parameters cannot be updated effectively during the late iterations. RMSProp (Root Mean Square Propagation), an improvement over AdaGrad, uses the Exponential Moving Average (EMA) to replace the sum of squared gradients \cite{zou2019sufficient}. RMSProp focuses on the fluctuations in recent gradients and alleviates the problem of learning rate decay inherent to AdaGrad. Kingma et al. proposed the Adam algorithm, which uses both the first and second moment estimates to correct the average gradient \cite{kingma2015adam}. Adam dynamically corrects the first and second moment estimates, and adjusts the learning rate for each parameter, which not only alleviates momentum underestimation but also further alleviates the issue of learning rate decay.

Muon \cite{jordan2024muon,grishina2025accelerating} was proposed after the boom of large language models (LLMs). This algorithm applies Newton-Schulz iteration to perform approximate orthogonalization on the momentum update matrix of SGD, thereby avoiding continuous learning rate decay, and significantly accelerating the training process of both image and language modeling tasks.

To sum up, gradient-based optimization algorithms in Euclidean space mainly focus on accurately representing gradient vectors and alleviating the issue of continuous learning rate decay. However, the geometry and geometric properties of objective functions are not considered.

\subsection{Gradient-based optimizers in Riemannian space}
Unlike Euclidean gradients, gradient-based optimizers for specialized tasks must incorporate manifold constraints. For example, this applies to tasks with geometric constraints on parameters/inputs \cite{bonev2025attention} and dedicated feature enhancement requirements \cite{fontana2025pose}.

Bonnabel firstly proposed a general gradient descent framework on Riemannian manifolds \cite{bonnabel2013stochastic}:
\begin{equation}
\label{equ_basic_GD_R}
  \boldsymbol{\theta_{t+1}} = exp_{\boldsymbol{\theta_t}}(-\eta H(\boldsymbol{x},\boldsymbol{\theta_t}))
\end{equation}
where $exp_{\boldsymbol{\theta_t}}$ is an exponential map at $\boldsymbol{\theta_t}$, and $H(\boldsymbol{x},\boldsymbol{\theta_t})$ denotes the Riemannian gradient of the objective function w.r.t $\boldsymbol{\theta_t}$. 

$H(\boldsymbol{x},\boldsymbol{\theta_t})$ can be obtained by \cite{qiu2025learning,fei2025survey}:
\begin{equation}
\label{equ_riemannian_gradient}
  H(\boldsymbol{x},\boldsymbol{\theta_t}) = \Pi_{T_{\boldsymbol{\theta_t}} \mathcal{M}} ( \nabla L(\boldsymbol{\theta_t};\boldsymbol{x_i})) 
\end{equation}
where $T_{\theta_t} \mathcal{M}$ denotes the tangent space at $\theta_t$, and $\Pi$ denotes an orthogonal projection.

\begin{figure}
  \centering
  \includegraphics[scale=0.65]{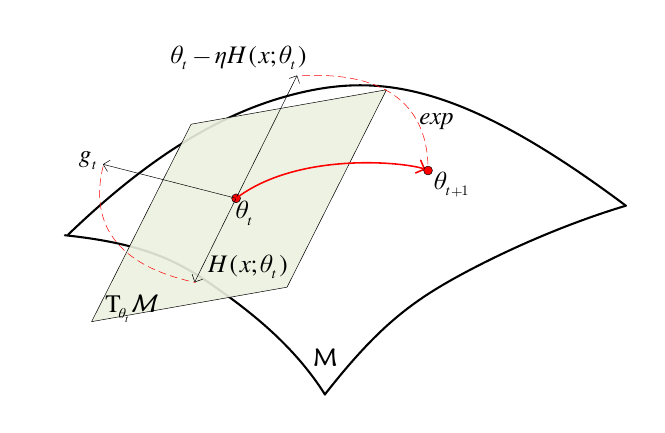}
  \caption{The Riemannian gradient \cite{fei2025survey}.}
  \label{fig_Riemannian_gradients}
\end{figure}

In Figure \ref{fig_Riemannian_gradients}, the Euclidean gradient $g_t$ is projected onto $T_{\theta_t} \mathcal{M}$ by an orthogonal projection to form the Riemannian gradient $H(\boldsymbol{x},\boldsymbol{\theta_t})$. The negative Riemannian gradient is then projected onto $\mathcal{M}$ via a retraction mapping to obtain a new parameter $\boldsymbol{\theta}_{t+1}$.

\begin{figure*}[t]
    \centering
    \subfloat[]{
       \includegraphics[scale=0.7]{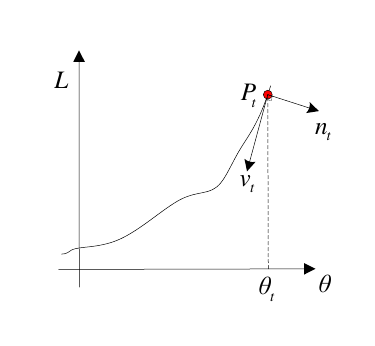}
    }
    \subfloat[]{
       \includegraphics[scale=0.7]{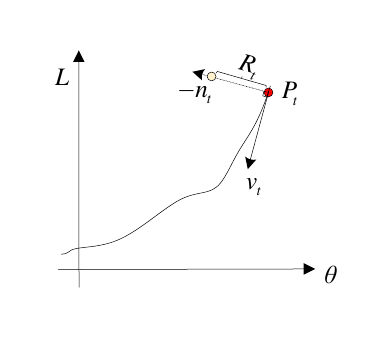}
    }
    \subfloat[]{
       \includegraphics[scale=0.7]{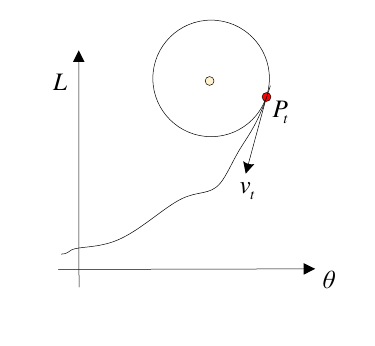}
    }\\
    \subfloat[]{
       \includegraphics[scale=0.7]{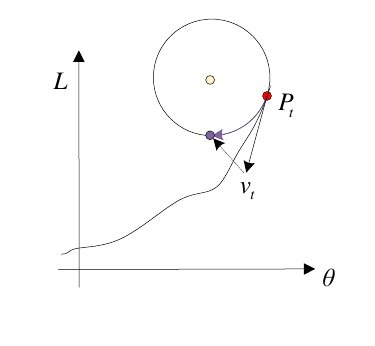}
    }
    \subfloat[]{
       \includegraphics[scale=0.7]{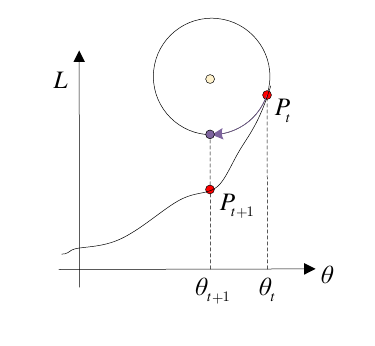}
    }
    \caption{The convergence process of GGD algorithm.}
    \label{fig_GGD_steps}
\end{figure*}

Based on the above framework, Shaikewitz et al. imposed constraints from the 3D Special Orthogonal Group (SO(3)) on the rotation matrix R to ensure its physical validity, thereby facilitating more accurate object pose estimation \cite{shaikewitz2025category}. Cheng et al. used the Grassmann manifold to construct a Riemannian Batch Normalization, which improves the accuracy in the field of manifold-based action recognition and signal classification on correlation matrix manifolds \cite{chen2025riemannian}. Bonev et al. incorporated 3D spherical constraints into the Transformer attention mechanism, enabling comprehensive global atmospheric observation for the earth \cite{bonev2025attention}. These studies are designed to exploit the intrinsic geometric structures of parameters or input data, which can significantly improve both the efficiency and accuracy of specialized tasks.

The objective function-induced hypersurfaces, the parameters and input data discussed in the above studies, are manifolds embedded in a high-dimensional Euclidean space. Different from parameters and input data, the complex geometry of the objective function-induced hypersurface is determined by neural network parameters, input data, and the objective function expression at the same time. Therefore, it may not be subject to explicit and intuitive physical or geometric constraints; even if such constraints exist, it is also difficult to represent this complex geometry using a single classic manifold constraint.

\section{METHODOLOGY}
\label{section_methodology}
\subsection{Overview}
Given a set of parameters $\boldsymbol{\theta}$, input data $\boldsymbol{x}$ and an objective function $L(\boldsymbol{x},\boldsymbol{\theta})$, the goal of GGD is to search for a parameter combination that minimizes $L(\boldsymbol{x},\boldsymbol{\theta})$ to its global optimum.

Considering the difficulty in visualizing the convergence progress of GGD in high-dimensional space, Figure \ref{fig_GGD_steps} illustrates the convergence process of GGD in a 2-D space (only one parameter requires updating). In each iteration, a point $P_t$ and the corresponding Euclidean gradient $\boldsymbol{g}$ can be calculated, see subgraph (a). The GGD first calculates the normal vector $\boldsymbol{n_t}$ from $\boldsymbol{g_t}$ at $P_t$, and then constructs an n-D sphere with radius $R_t$ tangent to the hypersurface at $P_t$ along $-\boldsymbol{n_t}$, see subgraphs (b) and (c). Second, $\boldsymbol{v_t}$ is projected onto the n-D sphere to form a geodesic with length equal to the norm of $\boldsymbol{v}_t$, see subgraph (d). Finally, a new point $P_{t+1}$ and a new set of parameters $\boldsymbol{\theta}_{t+1}$ can be obtained, see subgraph (e). 

\begin{algorithm}
\caption{The GGD algorithm.}
\label{alg_1}
\textbf{Require}: $L(\boldsymbol{x};\boldsymbol{\theta})$: the objective function with parameter $\boldsymbol{\theta} = \{\theta_1, \theta_2, \cdots, \theta_n\}$.\\
\textbf{Require}: $R_0$: the initial radius of the n-D sphere; $\sigma$: the width of an Gaussian function.\\
\begin{algorithmic}[1] 
\WHILE{$t <$  training iterations}
    \STATE $P_t \gets concat(\boldsymbol{\theta}_t, L(\boldsymbol{\theta}_t;\boldsymbol{x}))$\\
    \STATE $\boldsymbol{g}_t \gets \nabla_{\boldsymbol{\theta}_t} L(\boldsymbol{\theta}_t;\boldsymbol{x})$\\
    \STATE $\boldsymbol{n}_t \gets concat(\boldsymbol{g}_t,-1)$
    \STATE $\boldsymbol{v}_t \gets concat(\boldsymbol{g}_t, \| \boldsymbol{g}_t \|_2^2)$
    \STATE $R_t \gets R_0 \cdot exp(-0.5 \cdot (t-\mu)^2/\sigma^2)$\\
    \STATE $\boldsymbol{v}_t \gets 0.5 \cdot \pi R_t \boldsymbol{v}_t/\| \boldsymbol{v}_t \|$
    \STATE $C_t \gets R_t \boldsymbol{n}_t / ||\boldsymbol{n}_t||$
    \STATE $P_t \gets P_t-C_t$
    \STATE $P_{t+1} \gets cos(\frac{||\boldsymbol{v}_t||}{R_t})\cdot P_t+\frac{R_t \cdot sin(\frac{||\boldsymbol{v}_t||}{R_t})}{||\boldsymbol{v}_t||}\cdot \boldsymbol{v}_t$
    \STATE $P_{t+1} \gets P_{t+1}+C_t$
    \STATE $\boldsymbol{\theta}_{t+1} \gets P_{t+1}[0:n]$
\ENDWHILE
\STATE \textbf{return} $\boldsymbol{\theta}_{t+1}$ 
\end{algorithmic}
\end{algorithm}
\subsection{The calculation of the n-D sphere center}
Algorithm \ref{alg_1} shows the details of the GGD algorithm. In this section, we introduce lines 1-5 and 8-9 in the GGD algorithm.
 
Given a point $P_t=concat(\boldsymbol{\theta}_t, L(\boldsymbol{\theta}_t;\boldsymbol{x}))$ on the hypersurface induced by $L(\boldsymbol{\theta}_t;\boldsymbol{x})$, the gradient vector $\boldsymbol{g}_t$ at $P_t$ is $\boldsymbol{g}_t=\nabla_{\boldsymbol{\theta}_t} L(\boldsymbol{\theta}_t;\boldsymbol{x})$.

The normal vector and the tangent vector at $P_t$ can be calculated from $\boldsymbol{g}_t$:
\begin{equation}
\label{equ_gradient_vector}
\begin{cases}
  \boldsymbol{n}_t = (\frac{\partial L}{\partial \theta_1^t}, \frac{\partial L}{\partial \theta_2^t}, \cdots, \frac{\partial L}{\partial \theta_n^t}, -1)\\
  \boldsymbol{v}_t=concat(\boldsymbol{g}_t, \| \boldsymbol{g}_t \|_2^2).
\end{cases}  
\end{equation}
The detailed calculations of $\boldsymbol{n}_t$ and $\boldsymbol{v}_t$ can be found in Appendix \ref{appendix_vt}.

The center of the n-D sphere with radius $R_t$ can be calculated as:
\begin{equation}
\label{equ_center}
  C_t = P_t - R_t \frac{ \boldsymbol{n}_t}{||\boldsymbol{n}_t||}.
\end{equation}

Considering that the calculation of the geodesic is performed on an n-D sphere centered at the origin $O=\{0, 0, \cdots, 0\}$, it is therefore necessary to transport both $C_t$ and $P_t$ parallelly to an n-D sphere centered at the origin:
\begin{equation}
\label{equ_tranport}
\begin{cases}
  P_t = P_t-C_t\\
  C_t = O.
\end{cases}
\end{equation}
Because the vector $\boldsymbol{v}_t$ is calculated in Euclidean space, it does not need to be transported. In this section, we do not have to calculate the formula of the n-D sphere which is used to approximate a small neighborhood of the hypersurface induced by the objective function, we use the center $C_t$ and $R_t$ to calculate the projection of $\boldsymbol{v}_t$.

\subsection{Decay of $R_t$ and elimination of learning rate}
In this section, we introduce the lines 6-7 in the GGD algorithm.

\begin{figure}[h]
    \centering
    \includegraphics[scale=0.3]{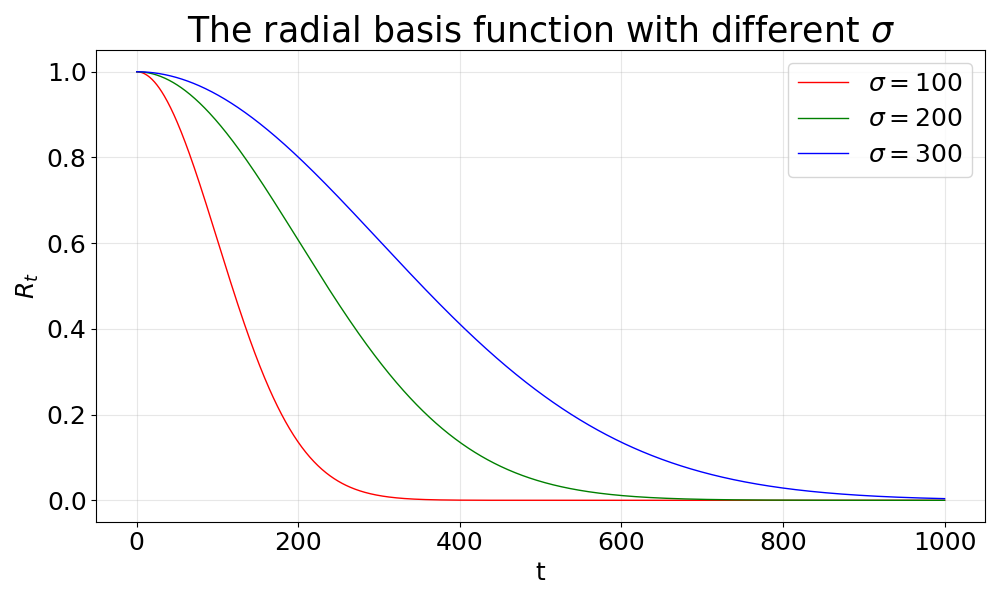}
    \caption{The variation curve of RBFs.}
    \label{fig_rbfs}
\end{figure}

As the number of iterations increases, the point $P_t$ gradually approaches the global minimum, and thus the radius $R_t$ should accordingly decay with the number of iterations. The decay of $R_t$ is:
\begin{equation}
\label{equ_decay_R}
  R_t = R_0 \cdot e^{- \frac{(t-\mu)^2}{2\sigma^2}}.
\end{equation}
where $R_t$ is actually a radial basis function (RBF) \cite{hu2022flow}, $R_0$ is the initial radius, $\mu$ and $\sigma$ are the center and width of the RBF, respectively. Since the variation curve of an RBF decreases gradually for $t>0$ (see Figure \ref{fig_rbfs}), which aligns with the decay of $R_t$, we select RBFs to control the radius decay.

\begin{figure}[h]
    \centering
    \includegraphics[scale=0.7]{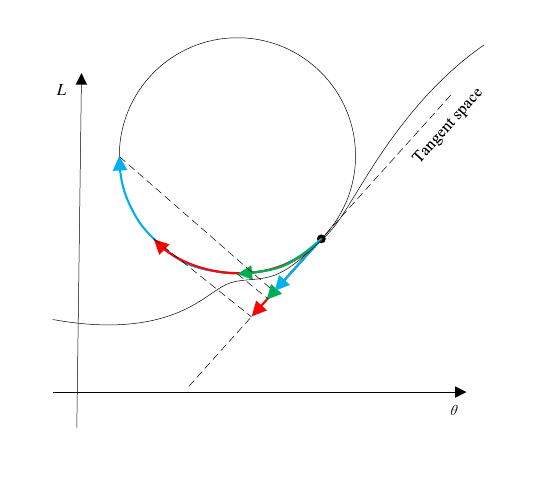}
    \caption{Different vectors in the tangent space of a circle and their corresponding projected geodesics on the circle.}
    \label{fig_quarter}
\end{figure}

Once the radius $R_t$ for the current iteration is determined, the tangent vector $\boldsymbol{v}_t$ can be scaled up or down. Because the length of  the projected geodesic equals the norm of the tangent vector (we will introduce this in section \ref{section_projection}),  the maximum norm of $\boldsymbol{v}_t$ is $\pi R_t/2$ (i.e., a quarter of the arc length). Therefore, the tangent vector $\boldsymbol{v}_t$ exhibits maximum length in the tangent space, as illustrated by the red vector in Figure \ref{fig_quarter}.

Therefore, the tangent vector $\boldsymbol{v}_t$ is scaled by:
\begin{equation}
\label{equ_max_vt}
  \boldsymbol{v}_t = 0.5 \cdot \pi R_t \frac{\boldsymbol{v}_t}{\| \boldsymbol{v}_t \|}.
\end{equation}

By scaling the tangent vector $\boldsymbol{v}_t$ up or down, we can approximate the maximum distance of movement along this direction within the tangent space.

\subsection{The projection of $\boldsymbol{v}_t$}
\label{section_projection}
In this section, we introduce lines 10-12 in the GGD algorithm.

The shortest distance between two points on a curved surface is determined by the geodesic with these two points as its start and end points \cite{mitra2026jacobi}. Therefore, we project the tangent vector $\boldsymbol{v}_t$ onto the n-D sphere to form a geodesic with length equal to $||\boldsymbol{v}_t||$, thus finding the correct convergence direction on the hypersurface.

The geodesic of an n-D sphere is on a circle centered at the center of the n-D sphere. The start point of the geodesic is $P_t$, and the length of the geodesic is $||\boldsymbol{v}_t||$. The function of geodesic is:
\begin{equation}
\label{equ_geodesic}
  \begin{cases}
    \gamma(s)=a(s)P_t+b(s)\boldsymbol{v}_t \\
    \gamma(0)=P_t\\
    \gamma'(0)=\boldsymbol{v}_t.
  \end{cases}
\end{equation} 
where $\gamma(s)$ denotes the geodesic and $s$ is the parameter of the geodesic. Because $\gamma(0)=P_t$ and $\gamma'(0)=\boldsymbol{v}_t$, the following formula can be obtained:
\begin{equation}
\label{equ_geodesic_1}
  \begin{cases}
    a(0)=1, b(0)=0\\
    a'(0)=0, b'(0)=1.
  \end{cases}
\end{equation}

The geodesic $\gamma(s)$ is used to describe the coordinate points on the geodesic on the n-D space, so $\gamma(s)$ can be expressed by a coordinate points on the geodesic:
\begin{equation}
\label{equ_geodesic_2}
  \gamma(s) = (x_1^s, x_2^s, x_3^s, \cdots, x_n^s).
\end{equation}  
Therefore, $||\gamma(s)||$ denotes the distance between $(x_1^s, x_2^s, x_3^s, \cdots, x_n^s)$ and the origin of the n-D space, i.e., $||\gamma(s)|| = R_t$, then we get:
\begin{equation}
\label{equ_geodesic_R}
  \begin{aligned}
  R_t =& a^2(s)P_t^2 + 2a(s)b(s)P_t\boldsymbol{v}_t+b^2(s)\boldsymbol{v}_t^2\\
    =& a^2(s)P_t^2 +b^2(s)\boldsymbol{v}_t^2\\
    =& a^2(s)||P_t||^2+b^2(s)||\boldsymbol{v}_t||^2.  
  \end{aligned}
\end{equation}
We remove ``$2a(s)b(s)P_t\boldsymbol{v}_t$'' because vector $P_t$ and $\boldsymbol{v}_t$ are perpendicular to each other.

Combining Equations (\ref{equ_geodesic_R}) and (\ref{equ_geodesic_1}), we can obtain the solution:
\begin{equation}
\label{equ_a_b}
  \begin{cases}
    a(s) = cos(\frac{||\boldsymbol{v}_ts||}{R_t})\\
    b(s) = \frac{R_t}{||\boldsymbol{v}_t||}sin(\frac{||\boldsymbol{v}_ts||}{R_t}).
  \end{cases}
\end{equation}
Then, the geodesic function can be expressed as:
\begin{equation}
\label{equ_a_b}
    \gamma(s) = cos(\frac{||\boldsymbol{v}_ts||}{R_t}) P_t+ \frac{R_t}{||\boldsymbol{v}_t||}sin(\frac{||\boldsymbol{v}_ts||}{R_t}) \boldsymbol{v}_t.
\end{equation}

We let $s=1$, and denote the end point of $\gamma(s)$ as $P_{t+1}$. Considering that we transport $P_t$ onto the origin-centered sphere in line 9, then we still need to transport $P_{t+1}$ back to the original sphere in line 11. The first $n$ components of the transported $P_{t+1}$ are the updated parameters $\boldsymbol{\theta}_{t+1}$.

\section{EXPERIMENTAL RESULTS AND ANALYSIS}
\label{section_experimental_result}

In this section, we compare 6 different gradient descent optimization algorithms (i.e., Adam \cite{kingma2015adam}, SGD \cite{deng2013recent}, SGDM \cite{sutskever2013importance}, Muon \cite{jordan2024muon}, SSGD \cite{sadrtdinov2025sgd} and our method GGD) on regression and classification tasks.

\subsection{Experiment I: Regression on Burgers' Flow Field Dataset}
The Burgers’ equation is a 1-D nonlinear partial differential equation (PDE) that expresses the movement of a shockwave across a tube:
\begin{equation}
\nonumber
\label{equ_burgers}
  \frac{\partial u}{\partial t}+u \frac{\partial u}{\partial x}=v \frac{\partial^2 u}{\partial x^2}
\end{equation}
where $u$ denotes the velocity of the fluid (the shockwave), $t$ denotes the time, $x$ denotes the displacement, and $v$ denotes the viscosity coefficient. 

\begin{table}[h]
    \centering
    \caption{The hyper parameters of the optimizers compared in this paper.}
    \label{tab_hyper_params}
    \begin{tabular}{cc}
        \hline
        Optimizer& Hyper parameters \\
        \hline
        SGD& learning rate=0.001\\
        SGDM&learning rate=0.001;momentum=0.9\\
        ADAM&learning rate=0.001;$\beta_1$=0.9;$\beta_2$=0.999\\
        Muon&learning rate=0.001;$\beta_1$=0.9\\
        SSGD&learning rate=0.001\\
        GGD&$\mu$=0; $\sigma$ and $R_0$ are adjusted.\\         
        \hline
    \end{tabular}
\end{table}

The variation range of the flow parameters ($t$, $x$, and $v$) is from 0.2 to 10.0, and the step size is 0.2. Consequently, this dataset contains 125,000 samples. In this dataset, $t$, $x$, and $v$ are the inputs of neural networks and $u$ is the output of the neural network.

\begin{table*}[h!]
    \centering
    \caption{The training loss of gradient descent optimizations on Burgers' dataset.}
    \label{tab_burgers}
    \begin{tabular}{cccccccccc}
        \hline
        Nets            & Structures           &MSE        & SGD\cite{deng2013recent}    &SGDM\cite{sutskever2013importance}    & Adam\cite{kingma2015adam}    & Muon\cite{jordan2024muon}   & SSGD\cite{sadrtdinov2025sgd} & GGD(ours)     & \makecell{reduction\\ (GGD V.S. Adam)} \\
        \hline
        \multirow{2}{*}{FCN\_1}&\multirow{2}{*}{3,6,1}&Training          &1.83E-3 &9.23E-3 & 1.37E-4 &4.60E-4 & 1.81E-3&\textbf{1.00E-4} & 27.01\%\\
                               &                      &Test &2.80E-3 &1.60E-3 & 4.47E-4 &1.07E-3 & 2.61E-3&\textbf{2.29E-4} & 48.76\%\\ 
        \multirow{2}{*}{FCN\_2}&\multirow{2}{*}{3,4,6,4,1}&Training      &1.47E-3 &6.40E-4 & \textbf{1.88E-5} &4.31E-5 & 2.27E-3&1.91E-5  &  -1.59\%\\
                               &                               &Test &2.04E-3 &1.20E-3 & 3.03E-4 &2.13E-4 & 3.07E-3&\textbf{1.80E-4}  &  40.59\%\\
        \multirow{2}{*}{FCN\_3}&\multirow{2}{*}{3,10*10,1 }&Training     &1.27E-3 &4.27E-4 & 2.00E-6 &1.73E-5 & 1.19E-3&\textbf{5.12E-7}  &  74.40\%\\            
                               &                               &Test &3.74E-3 &3.88E-3 & 8.83E-4 &2.34E-3 & 3.13E-3&\textbf{5.67E-4}  &  35.79\%\\          
        \hline
    \end{tabular}
\end{table*}

\begin{figure*}[h!]
    \centering
    \subfloat[]{
       \includegraphics[scale=0.38]{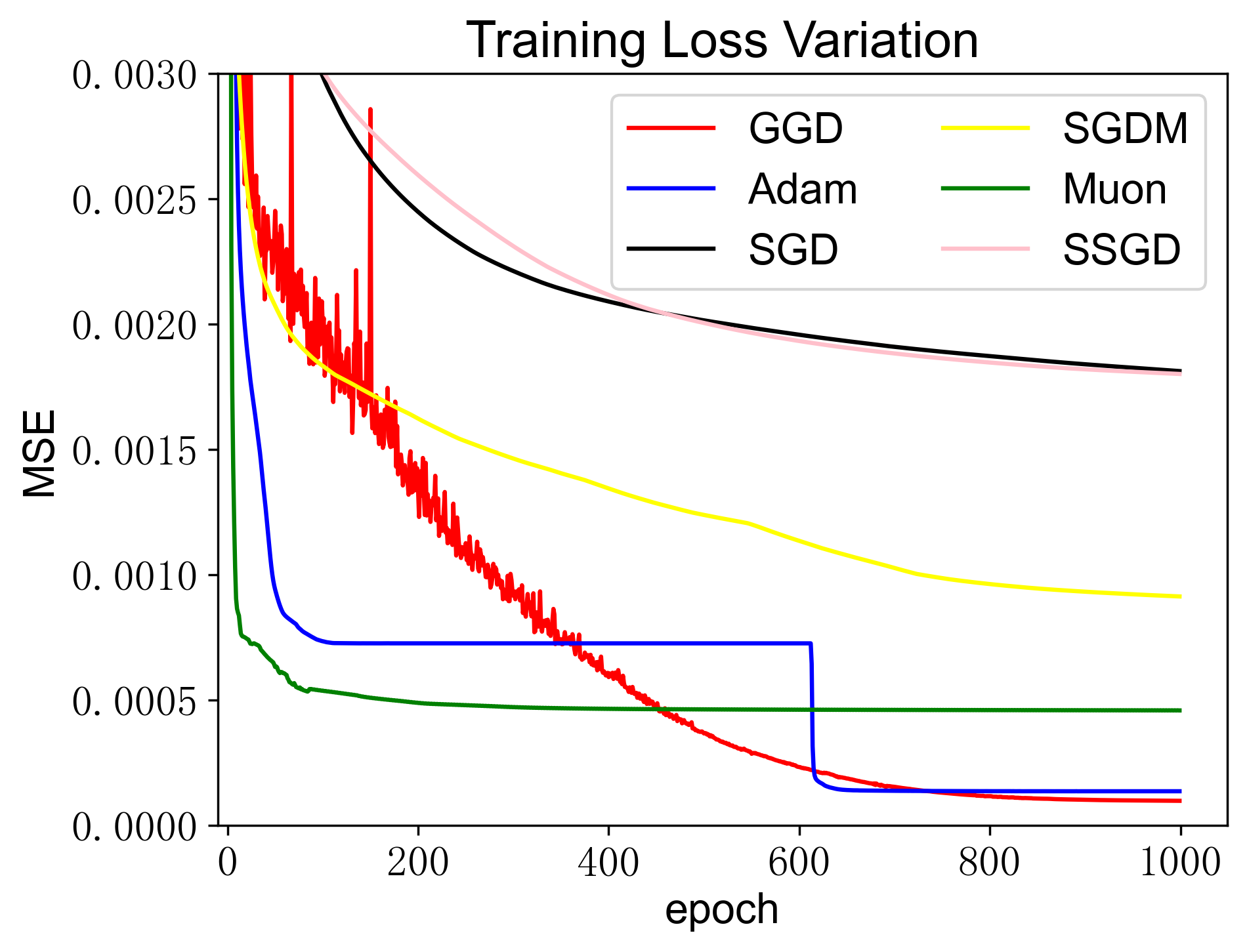}
    }
    \subfloat[]{
       \includegraphics[scale=0.38]{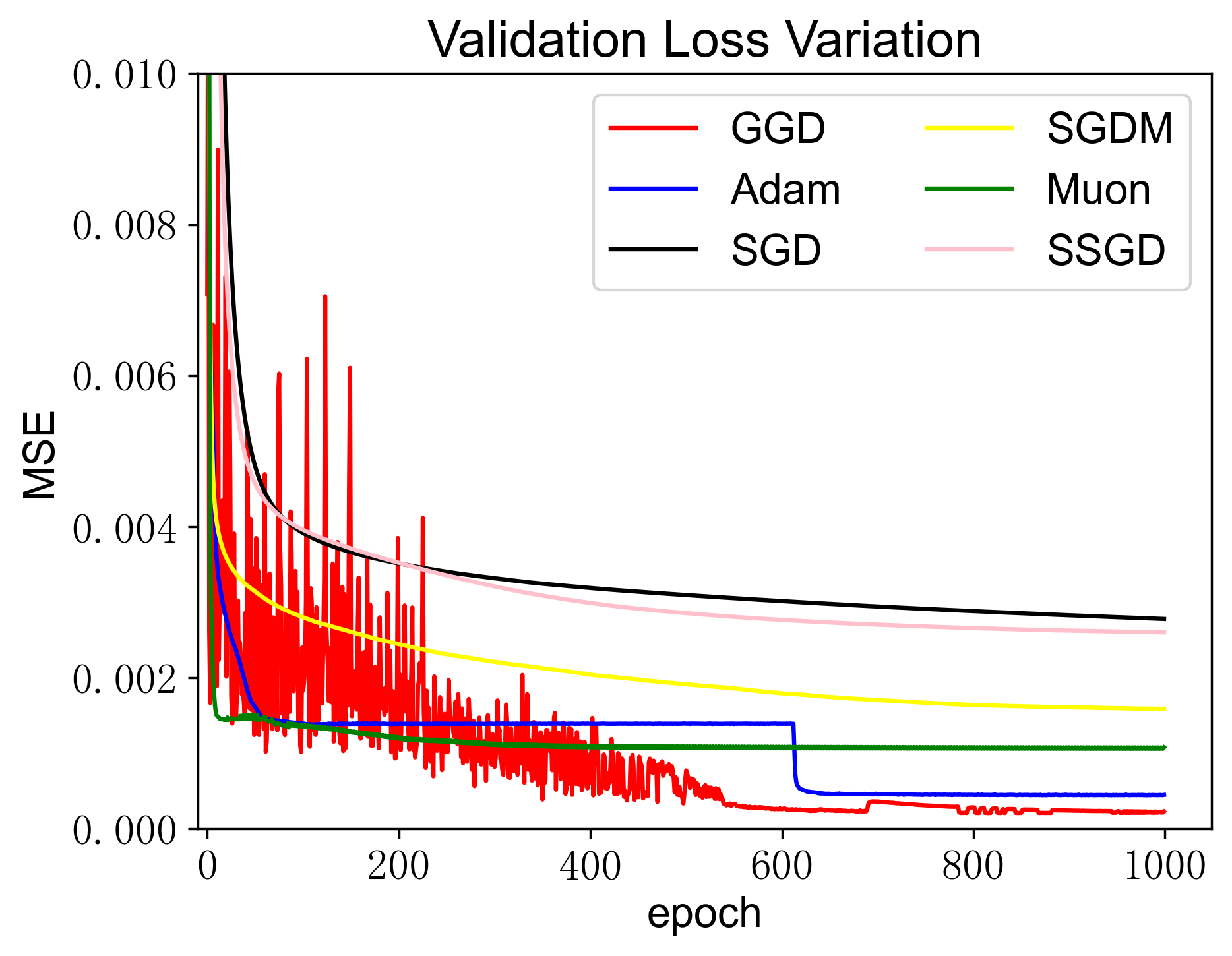}
    }\\
    \subfloat[]{
       \includegraphics[scale=0.38]{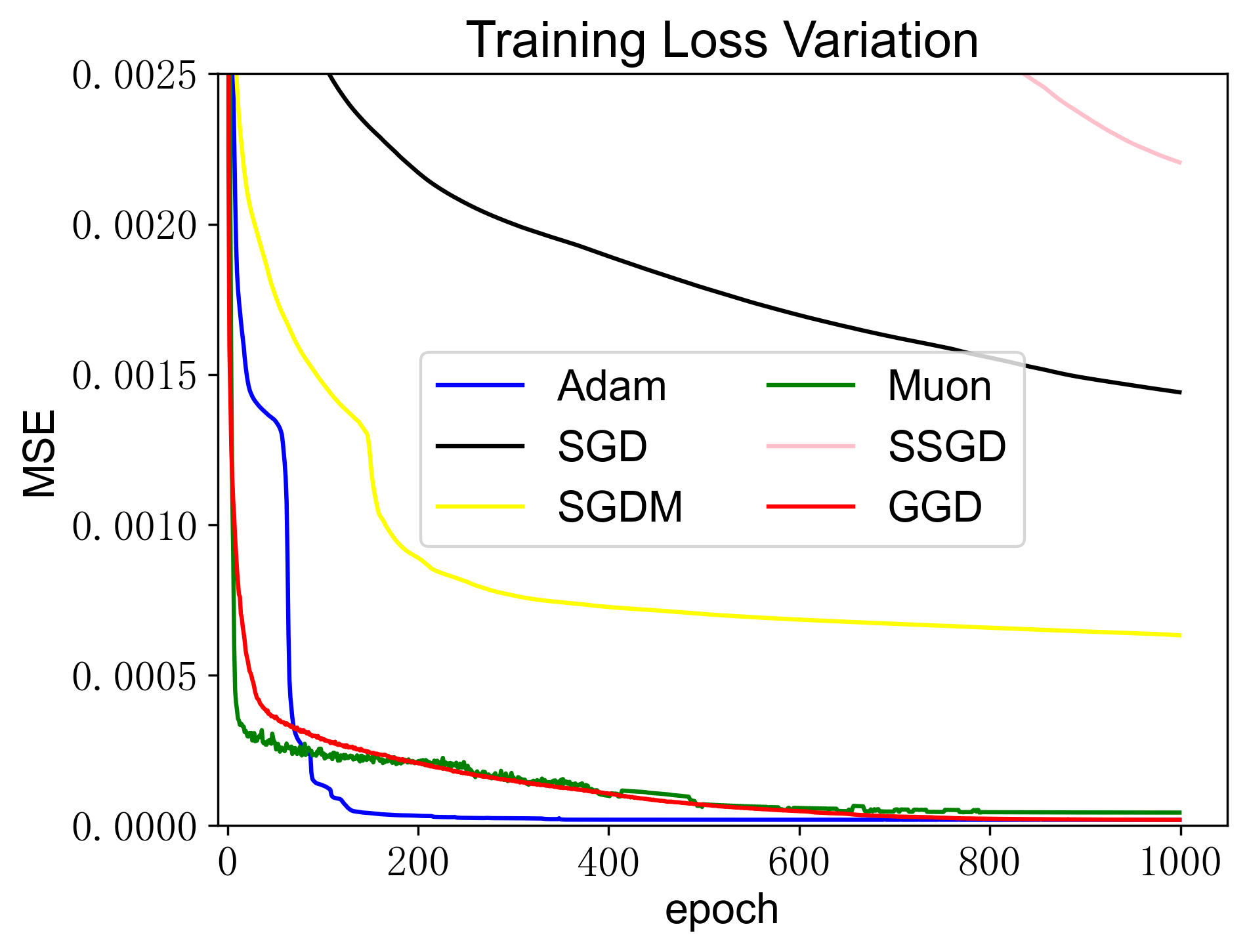}
    }
    \subfloat[]{
       \includegraphics[scale=0.38]{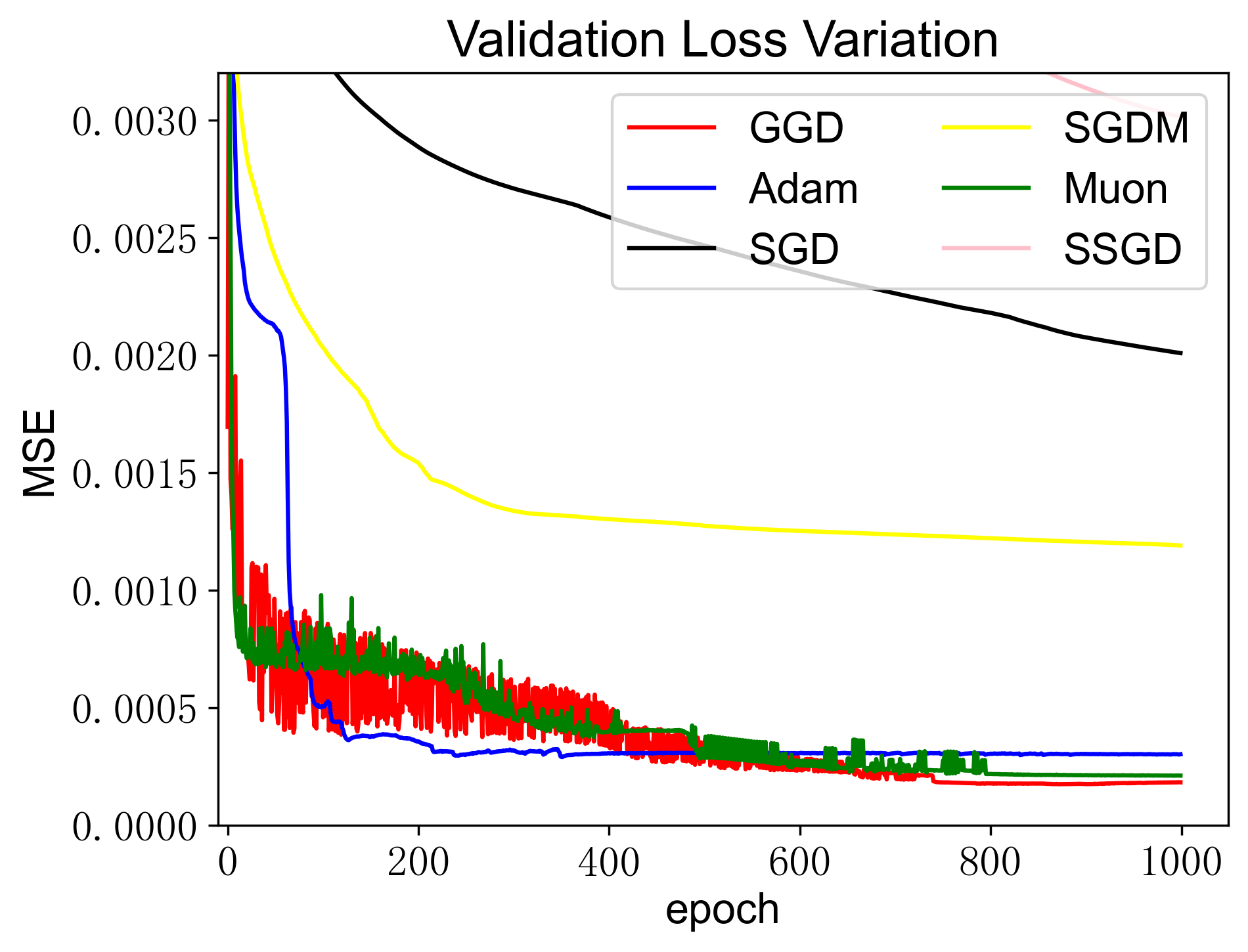}
    }\\
    \subfloat[]{
       \includegraphics[scale=0.38]{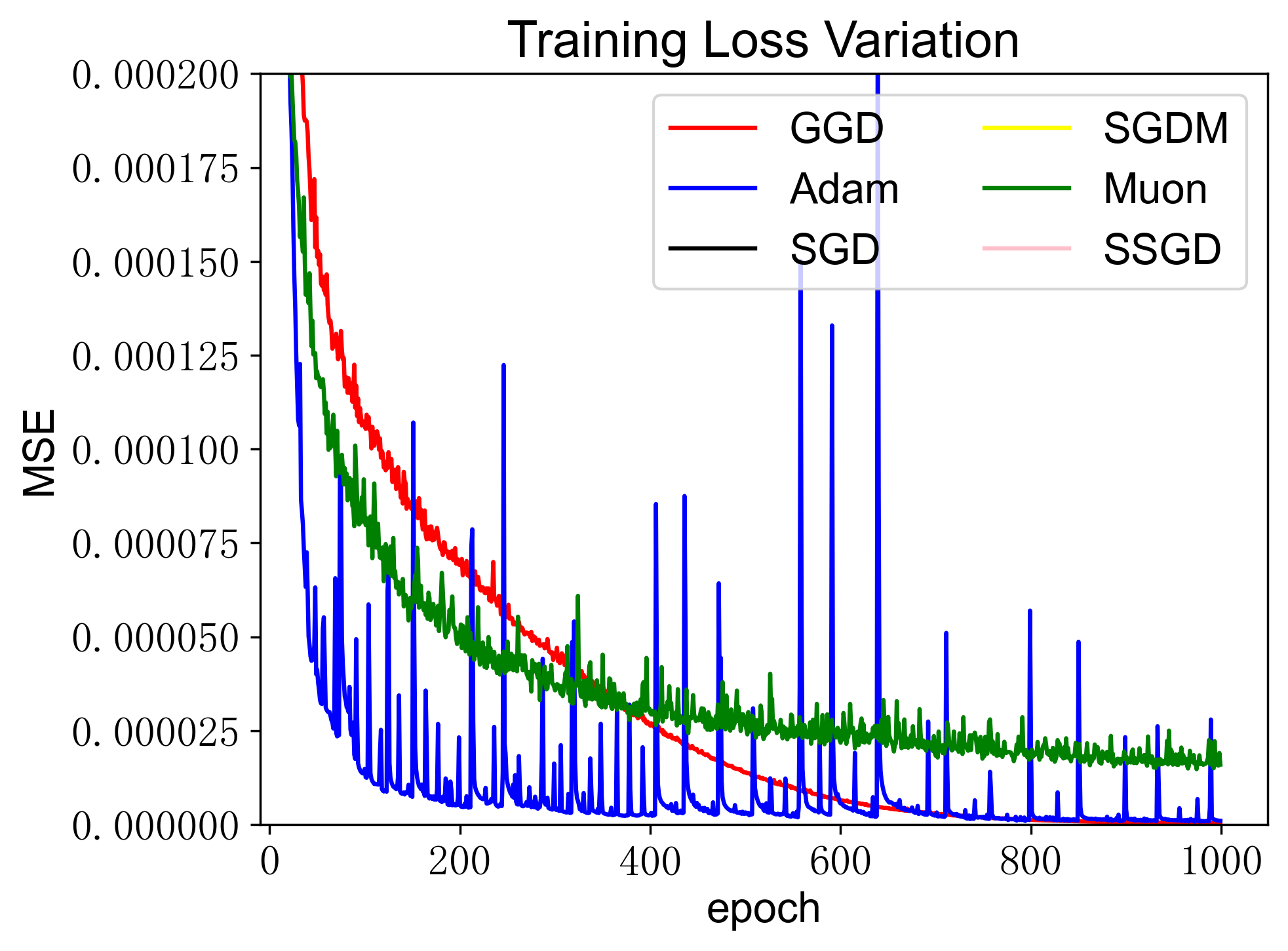}
    }
    \subfloat[]{
       \includegraphics[scale=0.38]{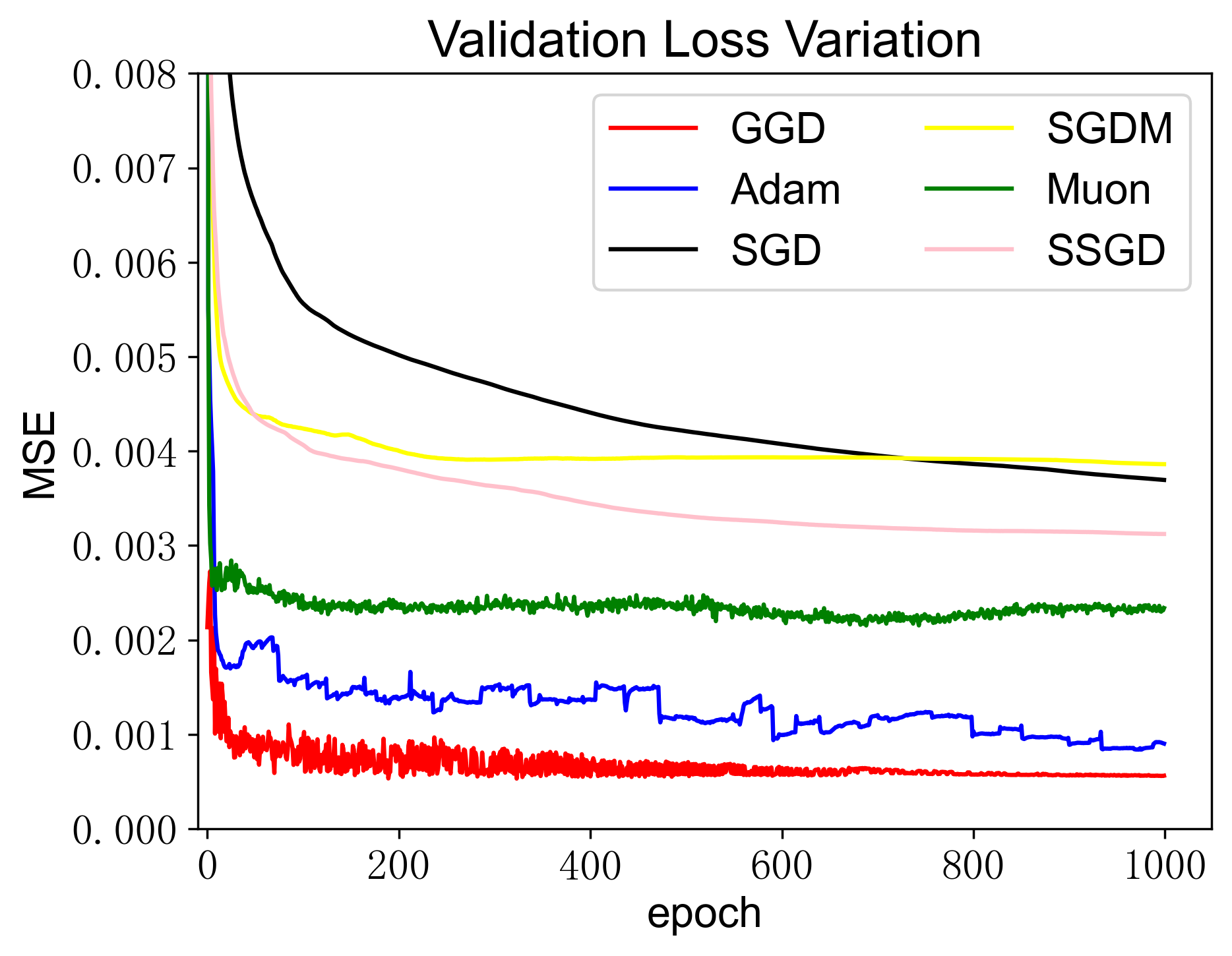}
    }
    \caption{The training and validation progress of all comparative algorithms on Burgers' dataset.}
    \label{fig_loss_burgers}
\end{figure*}

\begin{table*}[t]
    \centering
    \caption{The training loss of gradient descent optimizations on MNIST dataset.}
    \label{tab_mnist}
    \begin{tabular}{cccccccccc}
        \hline
        Nets& Structures        &Errors                         & SGD\cite{deng2013recent}    &SGDM\cite{sutskever2013importance}    & Adam\cite{kingma2015adam}    & Muon\cite{jordan2024muon}   & SSGD\cite{sadrtdinov2025sgd} & GGD(ours)     & \makecell{reduction\\ (GGD V.S. Adam)} \\
        \hline
        \multirow{3}{*}{CNN\_1}&\multirow{3}{*}{2*(3*3)+10}&Tr CE        &1.96E-01&6.72E-02&1.74E-02&3.94E-02&7.89E-01&\textbf{1.36E-2} &11.49\%\\
                               &                               &Te CE    &1.38E-01&4.87E-02&3.97E-02&3.79E-02&6.97E-01&\textbf{3.44E-02}&11.59\%\\
                               &                               &Acc      &95.92\% &98.34\% &98.85\% &98.71\% &89.12\% &\textbf{99.04\%}          &—\\   
        \multirow{3}{*}{CNN\_2}&\multirow{3}{*}{3*(3*3)+10}&Tr CE        &2.93E-01&8.56E-02&1.32E-02&2.32E-02&1.91    &\textbf{1.16E-02}&12.12\%\\
                               &                               &Te CE    &1.70E-01&5.00E-02&2.87E-02&3.11E-02&1.85    &\textbf{2.78E-02}&3.14\%\\
                               &                               &Acc      &95.20\% &98.48\% &99.14\% &99.00\% &63.30\% &\textbf{99.16\%}          &—\\
        \multirow{3}{*}{CNN\_3}&\multirow{3}{*}{3*(3*3)+128,10}&Tr CE    &3.50E-01&8.05E-02&7.48E-03&1.90E-02&2.18    &\textbf{3.38E-03}&28.07\%\\
                               &                               &Te CE    &2.07E-01&4.55E-02&3.22E-02&2.93E-02&2.16    &\textbf{2.83E-02}&6.21\%\\
                               &                               &Acc      &94.60\% &98.34\% & 99.20\%&99.07\% &24.27\% &\textbf{99.30\%}          &—\\          
        \hline
    \end{tabular}
\end{table*}
We designed three fully connected networks (FCNs) with different structures. The first FCN structure (FCN\_1) is denoted as ``3, 6, 1'', where the first number ``3'' and the last number ``1'' represent the number of neurons in the input layer and output layer, respectively. The middle number ``6'' indicates the number of neurons in the hidden layer. The second FCN structure (FCN\_2) is ``3, 4, 6, 4, 1''. Similarly, this FCN also has 3 neurons and 1 neuron in the input layer and output layer. The numbers ``4, 6, 4'' mean this FCN has three hidden layer, and each of which has 4, 6 and 4 neurons, respectively. The last FCN structure (FCN\_3) is ``3, 10*10, 1", which means this FCN has 10 hidden layers, each of which has 10 neurons. In addition, Table \ref{tab_hyper_params} shows the hyper parameters of the 6 gradient descent optimization algorithms.

Table \ref{tab_burgers} shows the training and test mean square errors (MSEs) of gradient descent optimizers, evaluated across three FCNs on the Burgers' dataset. As shown in this table, the proposed GGD algorithm outperforms SGD, SGDM, Muon and SSGD across the three FCNs in both training and test MSEs. Also, GGD show lower MSEs than those of Adam, except the test MSEs for FCN\_2. As a classic gradient descent algorithm, Adam generally achieves lower training and test MSEs than SGD and SGDM. SSGD is a non-Euclidean gradient descent algorithm that imposes spherical constraints on parameters, and the training and test MSEs of SSGD are higher than those of GGD. For FCN\_1, the proposed GGD achieves MSE reductions of 27.01\% (training) and 48.76\% (test) compared to Adam. For FCN\_2, GGD achieves respective MSE reductions of -1.59\% and 40.59\% compared to Adam in the training and test phases. In this case, the training MSE of GGD increases slightly, while the validation MSE of GGD is significantly reduced. Notably, for FCN\_3 with more hidden layers, GGD achieves a 74.40\% reduction in training MSE and a 35.75\% reduction in test MSE compared to Adam.

Figure \ref{fig_loss_burgers} shows the training and validation loss variations of the comparative algorithms during the training process. Subgraphs (a) and (b) are the training and validation loss variations on FCN\_1. Subgraphs (c) and (d) are the training and validation loss variations on FCN\_2. Subgraphs (e) and (f) are the training and validation loss variations on FCN\_3. As observed in subgraph (a), the training MSE of GGD is lower than that of Adam after convergence (i.e., $epoch> 800$). In subgraph (c), the training MSE of GGD is closer to that of Adam after convergence. And in subgraph (d), the variation curve of GGD exhibits slight fluctuations, whereas that of Adam shows severe fluctuations. Besides, the training MSE of GGD is lower than that of Adam after convergence. Subgraphs (b), (c) and (d) illustrate that the validation MSE of GGD remains the lowest among all comparative algorithms when $epoch >600$. In addition, it should be noted that the fluctuation of the validation MSEs of the GGD model decreases with the increase of network depth, indicating that the GGD model has a more stable training process in a more complex neural network structures.

In this experiment, the hyper parameters of GGD are: (1) For FCN\_1, $R_0=0.9$ and $\sigma=365$; (2) For FCN\_2, $R_0=0.21$ and $\sigma=369$; (3) For FCN\_3, $R_0=0.2$ and $\sigma=369$. How to chose the $R_0$ and $\sigma$ are introduced in \ref{section_hyper_parameters}.

\subsection{Experiment II: Classification on MNIST Images Dataset}
In this section, we validate the effectiveness of the GGD algorithm for CNN-based classification tasks on the MNIST dataset \cite{chauhan2024handwritten}.

We also designed three convolutional neural networks (CNNs) with different structures. The first CNN (CNN\_1), with the structure``2*(3*3)+10'', has two convolutional layers, and the kernel of each layer is $3 \times 3$. The outputs of convolutional layers are directly flattened into a 1-D tensor, and the size of the output tensor is 10. Similarly, the second CNN (CNN\_2), with the structure ``3*(3*3)+10'', has three convolutional layers, and the remaining settings are the same as those of the CNN\_1. The third CNN (CNN\_3), with the structure ``3*(3*3)+128,10'', adds a fully connected layer with 128 neurons, and the remaining settings of this structure are exactly the same as those of the CNN\_2.

Table \ref{tab_mnist} shows the training and test cross entropy (CE) values. From this table, we see that GGD achieves the smallest training and test CEs, as well as the highest accuracies, among all algorithms compared in this paper. Although the Adam algorithm achieves relatively low loss values and relatively high accuracies, GGD exhibits lower loss values and higher accuracies. The loss values of other algorithms (e.g., SGD, SGDM, SSGD and Muon) are relatively high, and the corresponding accuracies are smaller than those of Adam and GGD. For CNN\_1, GGD achieves CE reductions of 11.49\% (training) and 11.59\% (test) compared to Adam. For CNN\_2, GGD achieves respective CE reductions of 12.12\% and 3.14\% compared to Adam in the training and test phases. Notably, for the deeper CNN\_3, GGD achieves a 28.07\% reduction in training CE and a 6.21\% reduction in test CE compared to Adam. It is worth noting that the non-Euclidean gradient descent algorithm SSGD achieves the highest loss values and the lowest accuracies, because the hypersurface induced by the CE- based objective function is a complex manifold embedded in high-dimensional space whose constraint can not be represented by simple and single manifold constrain (e.g., a sphere constraint).

\begin{figure*}[t]
    \centering
    \subfloat[]{
       \includegraphics[scale=0.38]{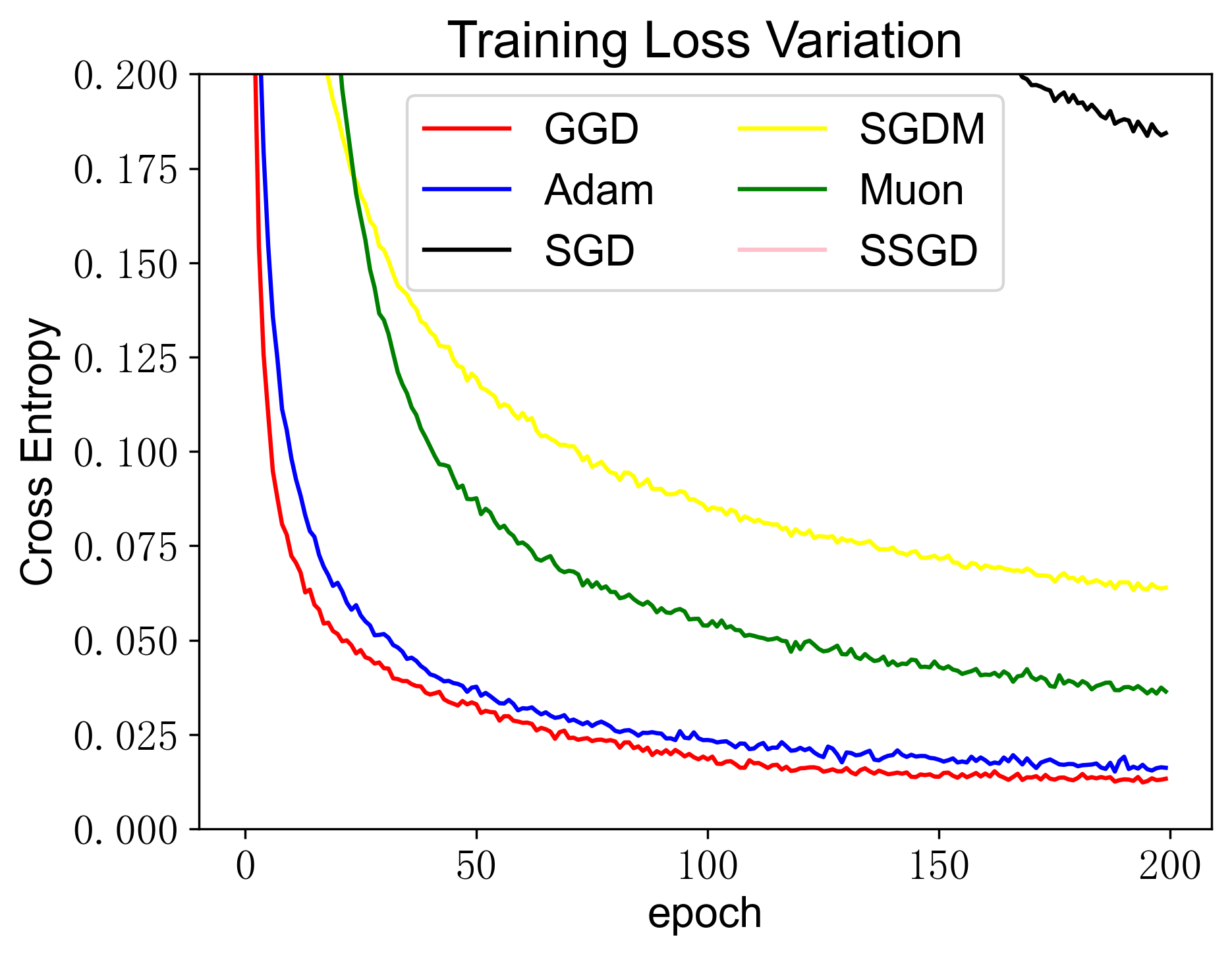}
    }
    \subfloat[]{
       \includegraphics[scale=0.38]{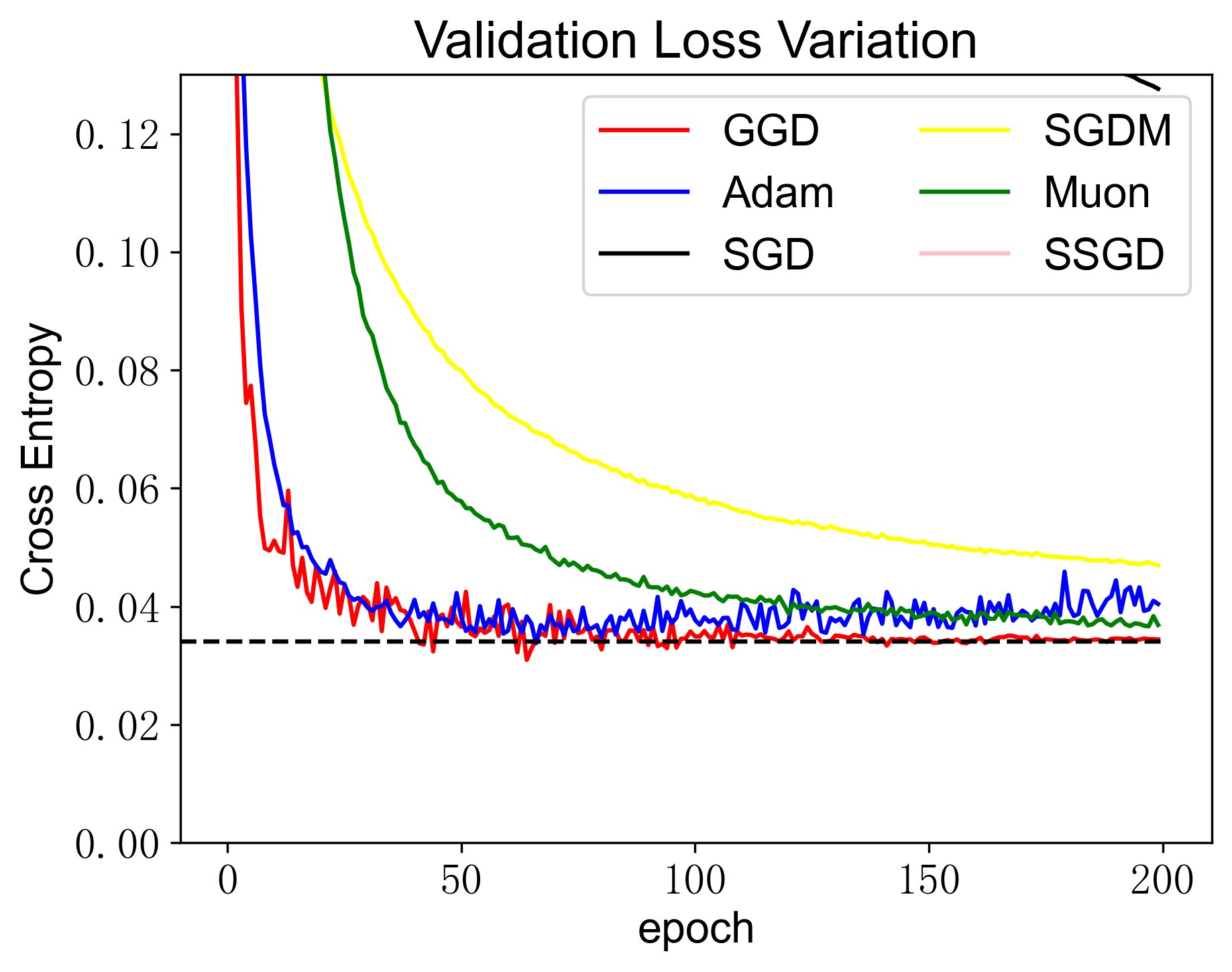}
    }\\
    \subfloat[]{
       \includegraphics[scale=0.38]{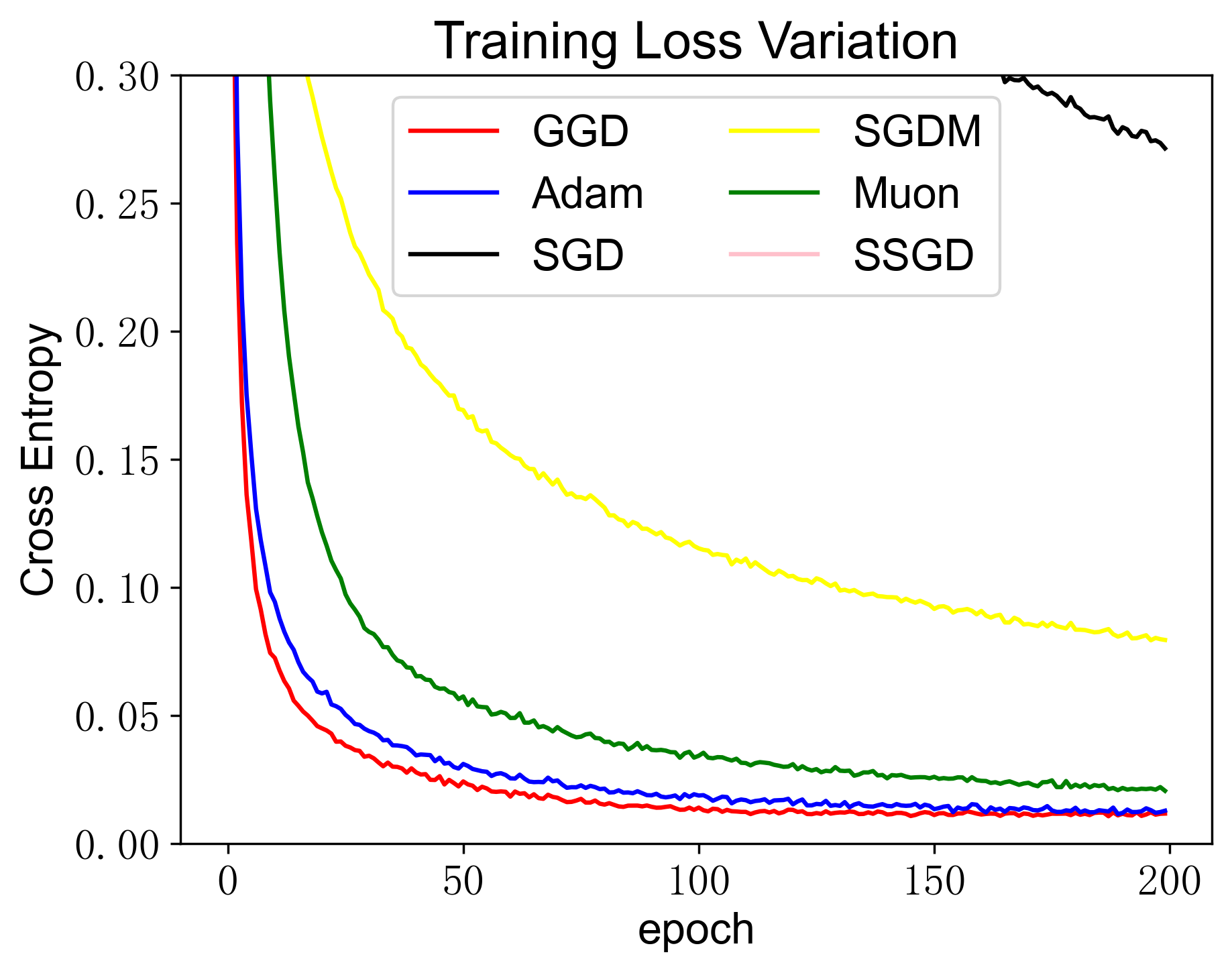}
    }
    \subfloat[]{
       \includegraphics[scale=0.38]{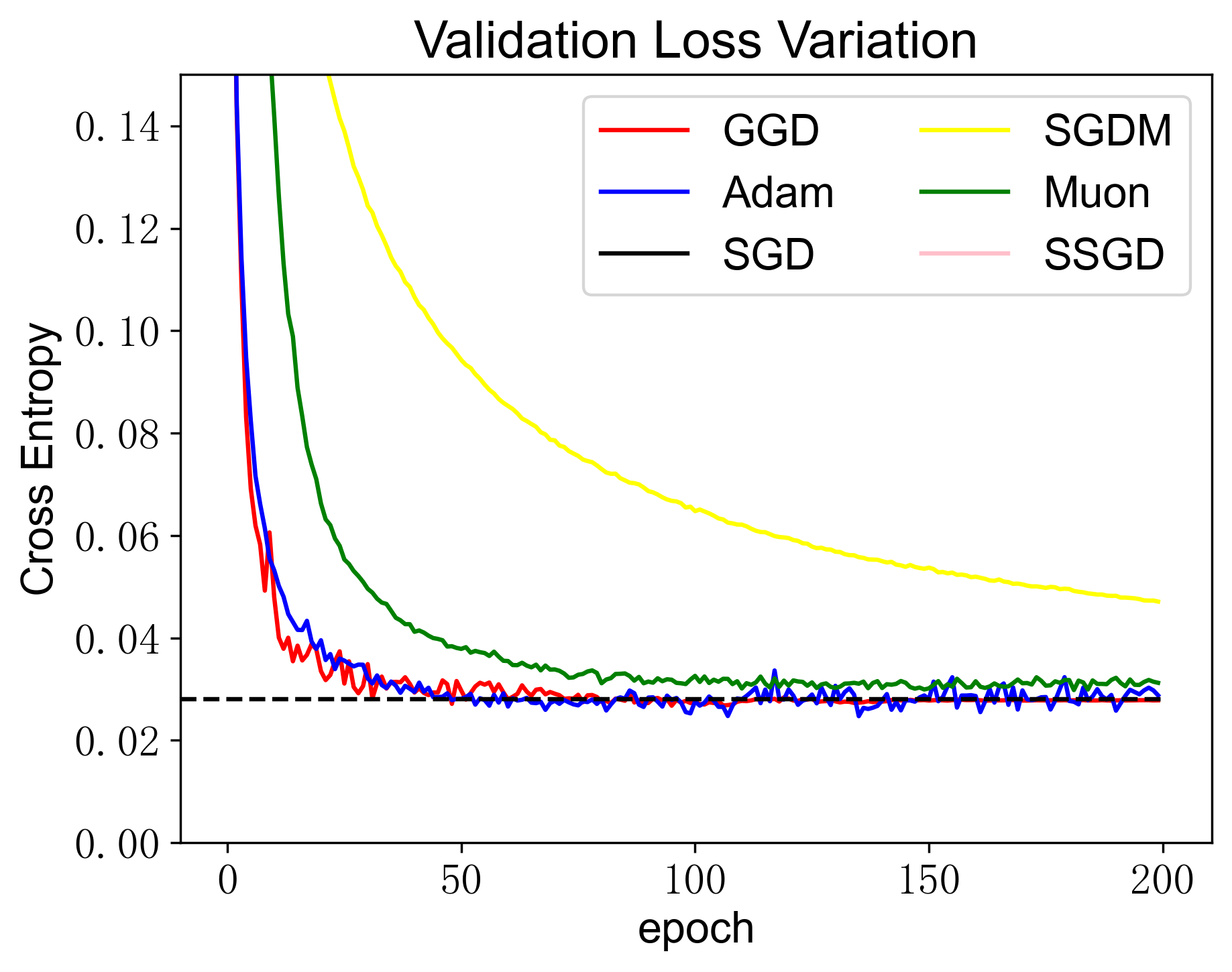}
    }\\
    \subfloat[]{
       \includegraphics[scale=0.38]{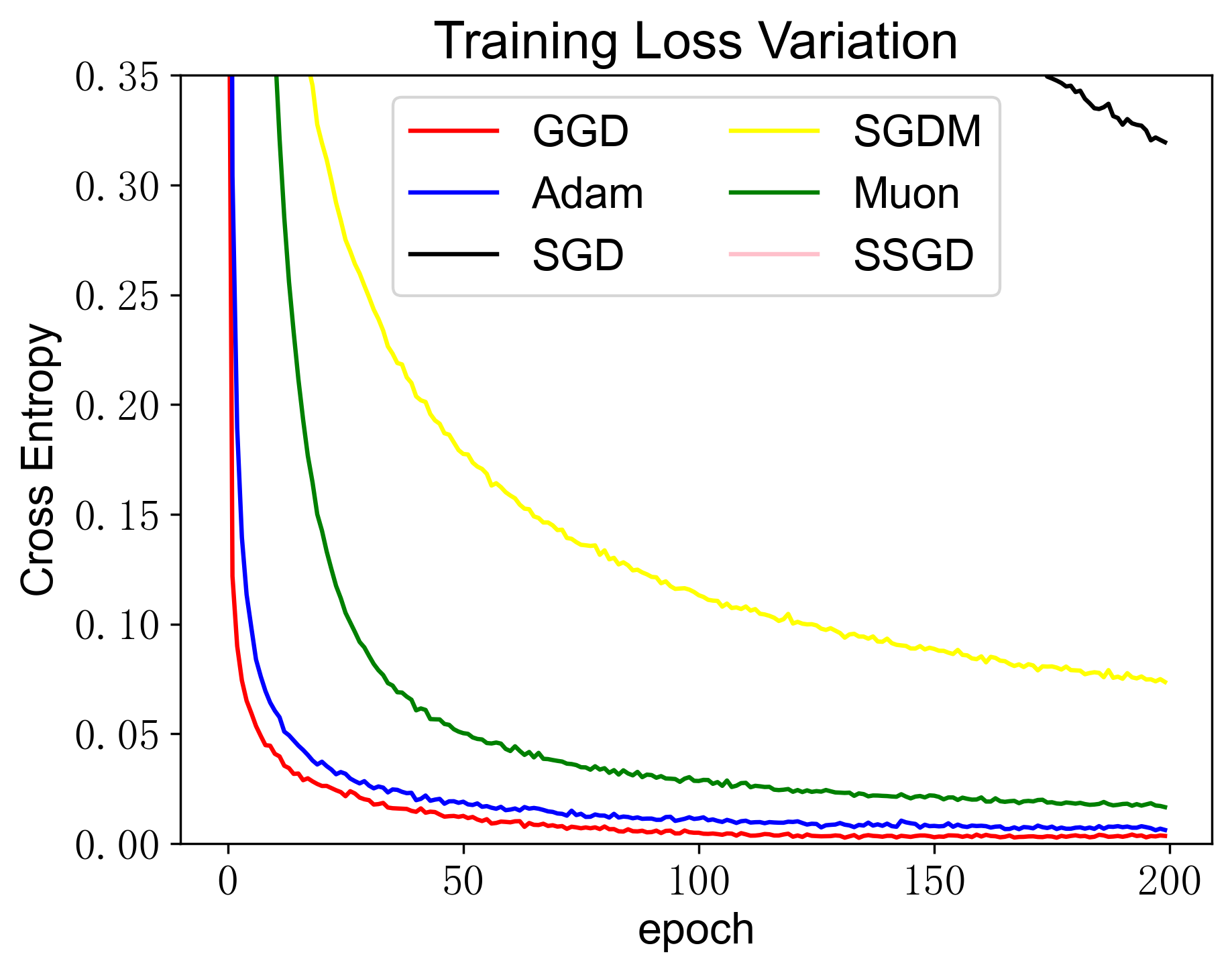}
    }
    \subfloat[]{
       \includegraphics[scale=0.38]{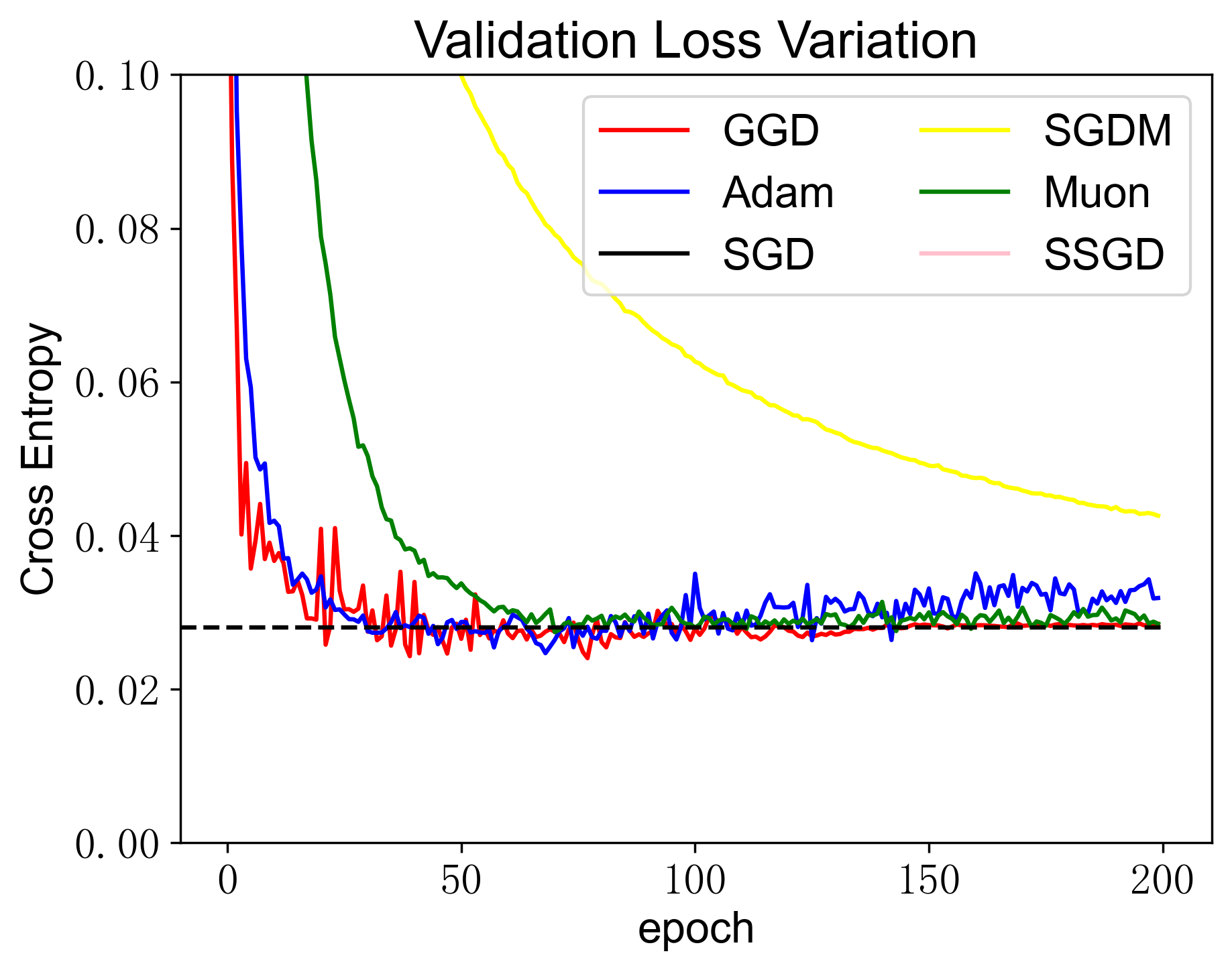}
    }
    \caption{The training and validation progress of all comparative algorithms on MNIST dataset.}
    \label{fig_loss_mnist}
\end{figure*}

Figure \ref{fig_loss_mnist} illustrates the training and validation CE loss variations of all comparative algorithms on the MNIST dataset. Subgraphs (a) and (b) are the loss variations on CNN\_1, subgraphs (c) and (d) are the loss variations of algorithms on CNN\_2, and subgraphs (e) and (f) are the loss variations of algorithms on CNN\_3. From subgraphs (a), (c) and (e), we see that the training loss values of GGD consistently remain the lowest. And in subgraphs (b) and (f), the loss curve of the Adam algorithm gradually rises after 100 iterations, which indicates the overfitting phenomenon. The loss variations of SGD and SSGD are omitted in some subplots, as their loss values exceed the vertical axis limit.

\begin{figure*}[t]
    \centering
    \subfloat[]{
       \includegraphics[scale=0.35]{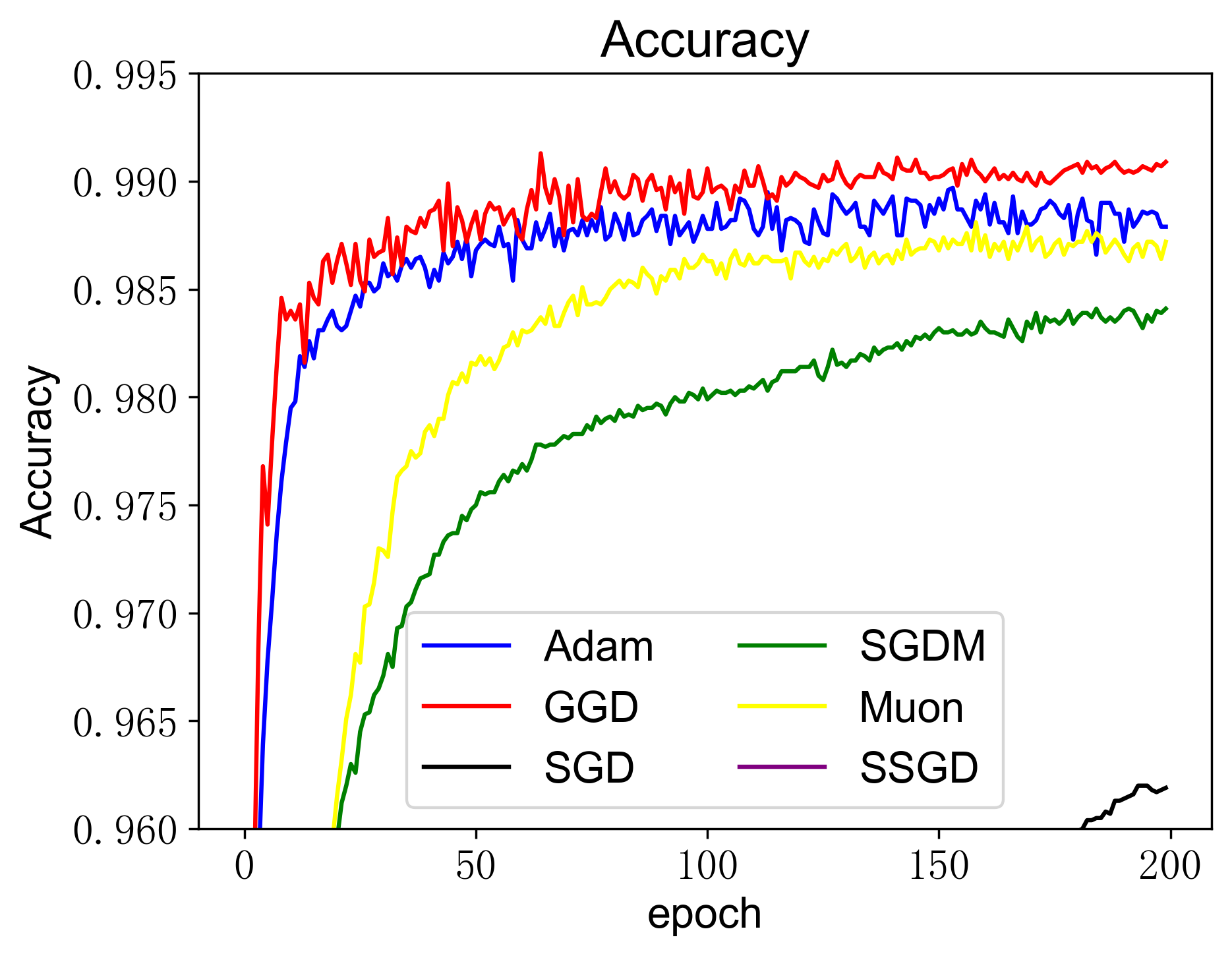}
    }
    \subfloat[]{
       \includegraphics[scale=0.35]{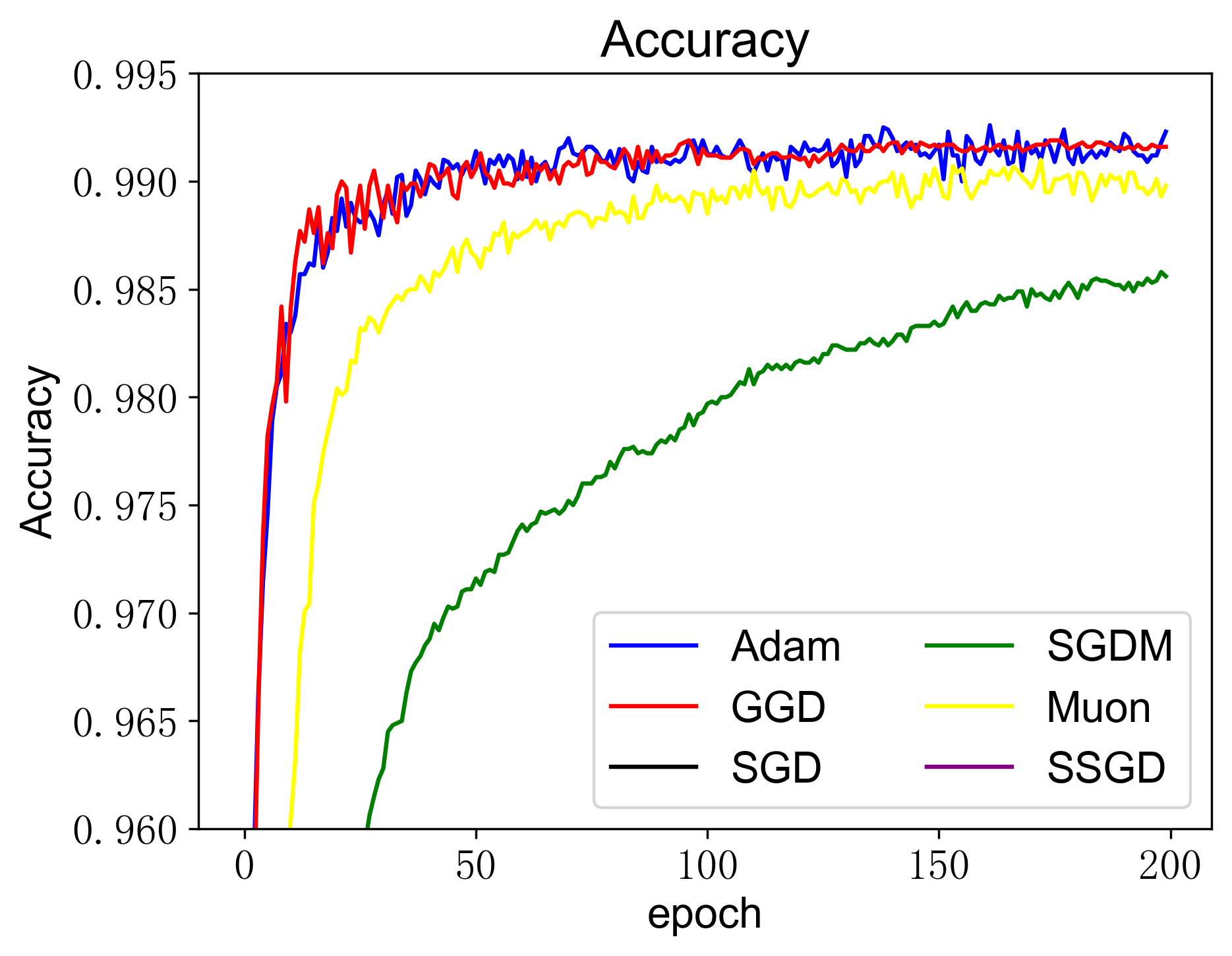}
    }
    \subfloat[]{
       \includegraphics[scale=0.35]{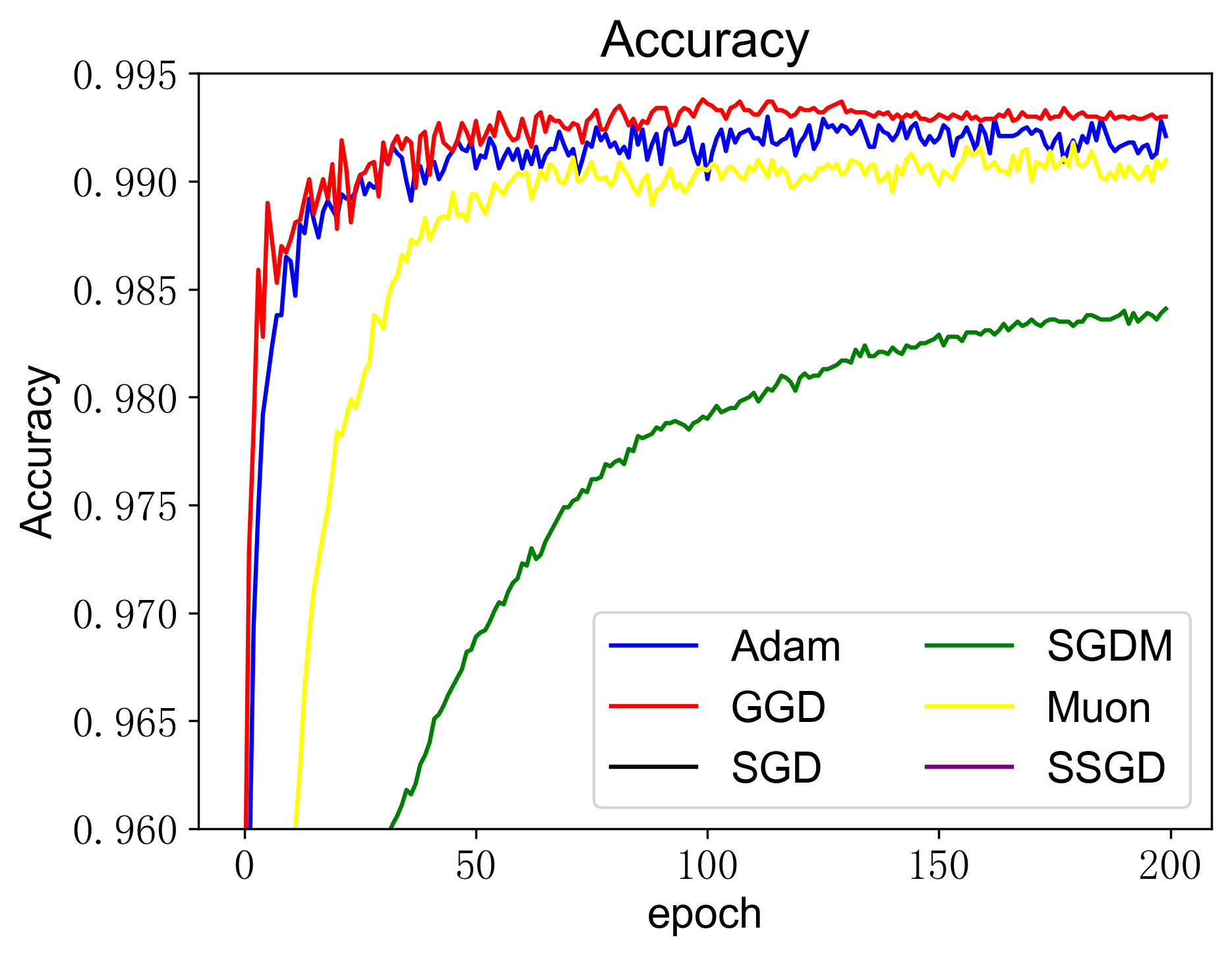}
    }
    \caption{The accuracy validation progress of all comparative algorithms on MNIST dataset.}
    \label{fig_acc}
\end{figure*}

Subgraphs (a), (b) and (c) in Figure \ref{fig_acc} show the accuracy variations of all comparative algorithms on CNN\_1, CNN\_2 and CNN\_3 during the training processes, respectively. In subgraph (a), it is evident that the GGD algorithm achieves the highest classification accuracy for MNIST handwritten digits. In subgraph (b), the accuracy variation of GGD algorithm is similar to that of the Adam. In subgraph (c), the accuracy variation of GGD exceeds that of the Adam after 50 iterations. The accuracy curves of remaining algorithms are lower than those of GGD. 

In this experiment, the hyper parameters of GGD are: (1) For CNN\_1, $R_0=0.5$ and $\sigma=65$; (2) For FCN\_2, $R_0=0.2$ and $\sigma=52$; (3) For FCN\_3, $R_0=1.0$ and $\sigma=50$. How to chose the $R_0$ and $\sigma$ are introduced in \ref{section_hyper_parameters}.

\subsection{Experimental Analysis} 

\subsubsection{Training time of the 6 gradient descent algorithms}
\begin{table*}[h]
    \centering
    \caption{The training time (seconds) of the 6 algorithms.}
    \label{tab_training_time}
    \begin{tabular}{ccccccccc}
        \hline
        Experiment               &  iterations          &Networks       & SGD    &SGDM    & Adam    & Muon   & SSGD & GGD\\
        \hline
        \multirow{3}{*}{Burgers'}&\multirow{3}{*}{1000} & FCN\_1        & 102.59 & 123.56 & 147.95  & 127.15 & 182.39 &161.30\\
                                 &                      & FCN\_2        & 179.99 & 181.81 & 265.88  & 242.45 & 271.87 &238.66\\
                                 &                      & FCN\_3        & 501.80 & 606.11 & 801.47  & 727.92 & 832.16 &697.33\\
        \multirow{3}{*}{MNIST}   &\multirow{3}{*}{200}  &CNN\_1         & 3258.90& 3220.50& 3409.10 & 3692.38& 3625.72&3468.02\\
                                 &                      &CNN\_2         & 3613.33& 3799.60& 3872.47 & 3667.76& 3677.74&3668.26\\
                                 &                      &CNN\_3         & 3615.88& 3631.42& 3788.23 & 3437.67& 3633.23&3617.78\\
        \hline
    \end{tabular}
\end{table*}

\begin{figure*}[h]
    \centering
    \subfloat[]{
       \includegraphics[scale=0.25]{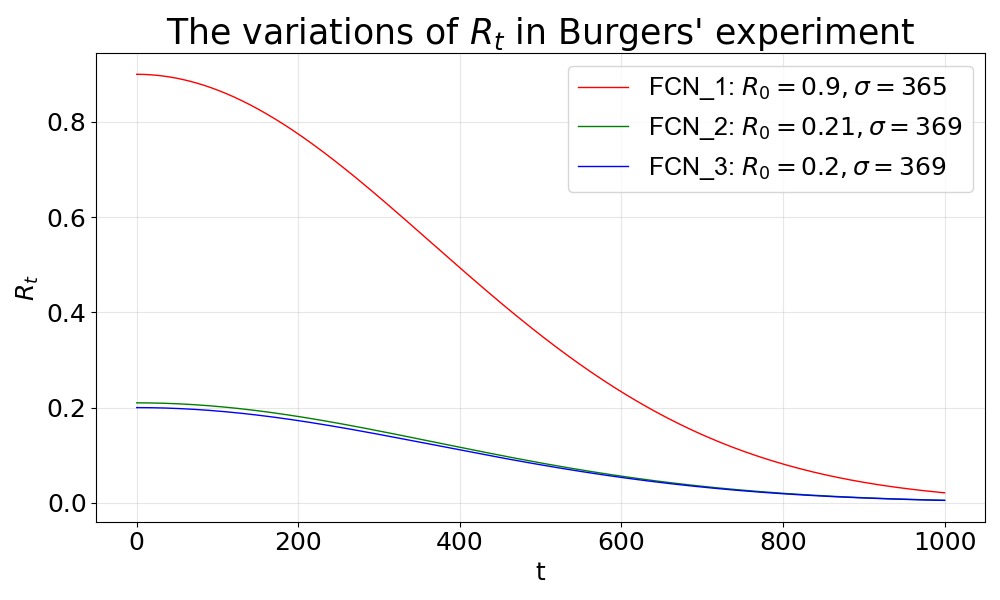}
    }
    \subfloat[]{
       \includegraphics[scale=0.25]{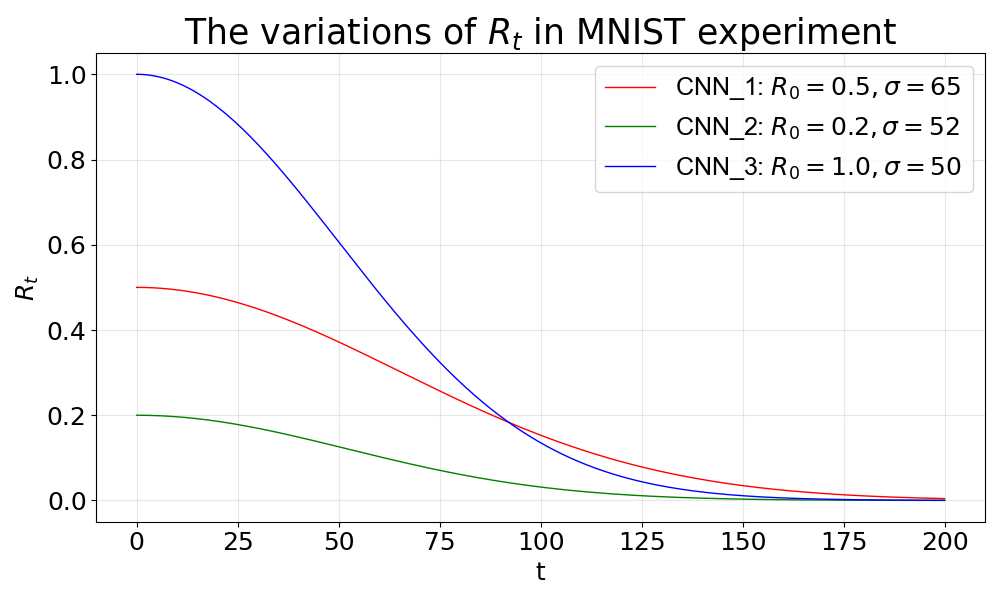}
    }
    \caption{The variations of $R_t$ in Burgers' and MNIST experiments.}
    \label{fig_R}
\end{figure*}

Table \ref{tab_training_time} presents the training time of 6 gradient descent algorithms on the two aforementioned experiments. In the Burgers’ experiment, the neural network FCN\_1 has only one hidden layer, and the training time of GGD is shorter than that of SSGD. For FCN\_2 with three hidden layers, GGD exhibits a lower training time than Adam, Muon and SSGD. When it comes to FCN\_3 with 10 hidden layers, the training time of GGD is also shorter than that of SSGD, Muon, and Adam. These results demonstrate that compared with other algorithms, the training speed of GGD accelerates as the number of hidden layers increases. A consistent trend is also observed in the MNIST experiment. For CNN\_1 with only 2 convolutional layers, the training time of GGD is shorter than that of SSGD and Muon. For CNN\_2 with 3 convolutional layers, GGD requires less training time than SSGD, Adam, and SGDM. For CNN\_3 with 3 convolutional layers and 1 fully connected layer, the training time of GGD is shorter the that of SGDM, Adam and SSGD.

\subsubsection{How to chose $R_0$ and $\sigma$}
\label{section_hyper_parameters}
The purpose of setting hyper parameters $R_0$ and $\sigma$ is to enable $R_t$ to decay gradually over the pre-configured iteration. Figure \ref{fig_R} depictes the variation trend of $R_t$ in both the Burgers' and MNIST experiments. The variations of $R_t$ is consistent with that of the radial basis functions ($t>0$). At the time of paper completion, the radius decay can only be controlled by hyper parameters. In future research, the radius decay may be directly derived from the curvature of the hypersurface, which can be represented by the Euclidean gradient and the power of the Euclidean gradient (i.e., a deterministic gradient descent algorithm without hyper parameters).

\section{CONCLUSION}
\label{section_conclusion}
We propose a generic, learning-rate-free optimizer designed for objective function-induced manifolds, namely the geodesic gradient descent algorithm. In this paper, an n-dimensional sphere is employed to approximate a small open neighborhood on the manifold. Generally, other types of structures (e.g., hyperbolic and Bézier surface) can also be utilized for this approximation, with the prerequisite of deriving their geodesic equations. In addition, since we use an n-dimensional sphere with radius $R_t$ to approximate the small open neighborhood on the manifold, the maximum step size is equivalent to a quarter of the arc length on the sphere. Consequently, we adopt this maximum step size for parameter updating instead of relying on a learning rate. Experimental results from regression and classification tasks demonstrate that GGD achieves lower training and test errors compared to existing algorithms.

However, the explicit determination of hyper parameters $R_0$ and $\sigma$ remains unclear. These parameters may be associated with the curvature of the manifold induced by the objective function, which may transform the gradient descent algorithm into a deterministic approach without the need of  hyper parameters.

\section*{Acknowledgments}
The authors would like to thank Prof. Stefan Sommer from the University of Copenhagen for providing computational assistance in geodesics.

{\appendix[The calculation of $\boldsymbol{n}_t$ and $\boldsymbol{v}_t$]
\label{appendix_vt}
Translate the given objective function $L = L(\boldsymbol{\theta}_t;\boldsymbol{x}) = f(\theta_1^t,\theta_2^t, \cdots, \theta_n^t; \boldsymbol{x})$ into implicit form:
\begin{equation}
\label{equ_loss_implicit}
    F = f(\theta_1^t,\theta_2^t, \cdots, \theta_n^t;\boldsymbol{x}) - L = 0.
\end{equation}

An objective function defined with $n$ parameters (i.e., $\boldsymbol{\theta}_t$) can be formulated as an implicit function defined with $n+1$ parameters (i.e., $\boldsymbol{\theta}_t$ and $L$).

The gradient vector of $F$ w.r.t $n+1$ parameters can be formulated
\begin{equation}
\label{equ_gradient}
\begin{aligned}
\nabla F =& (\frac{\partial F}{\partial \theta_1^t}, \frac{\partial F}{\partial \theta_2^t}, \cdots, \frac{\partial F}{\partial \theta_n^t}, -1)\\
         =&(\frac{\partial L}{\partial \theta_1^t}, \frac{\partial L}{\partial \theta_2^t}, \cdots, \frac{\partial L}{\partial \theta_n^t}, -1)
\end{aligned}
\end{equation}

The gradient vector is actually the normal vector, therefore, we get:
\begin{equation}
\boldsymbol{n}_t = \nabla F = (\frac{\partial L}{\partial \theta_1^t}, \frac{\partial L}{\partial \theta_2^t}, \cdots, \frac{\partial L}{\partial \theta_n^t}, -1).
\end{equation}

The tangent vector $\boldsymbol{v}_t$ is perpendicular to the $\boldsymbol{n}_t$, thus we get:
\begin{equation}
\begin{aligned}
\boldsymbol{v}_t = &(\frac{\partial L}{\partial \theta_1^t}, \frac{\partial L}{\partial \theta_2^t}, \cdots, \frac{\partial L}{\partial \theta_n^t}, \sum_{i}(\frac{\partial L}{\partial \theta_i^t})^2)\\
                 = & concat(\boldsymbol{g}_t, \| \boldsymbol{g}_t \|_2^2).
\end{aligned}
\end{equation}}


\bibliographystyle{IEEEtran}
\bibliography{main}

\begin{IEEEbiography}[{\includegraphics[width=1in,height=1.25in,clip,keepaspectratio]{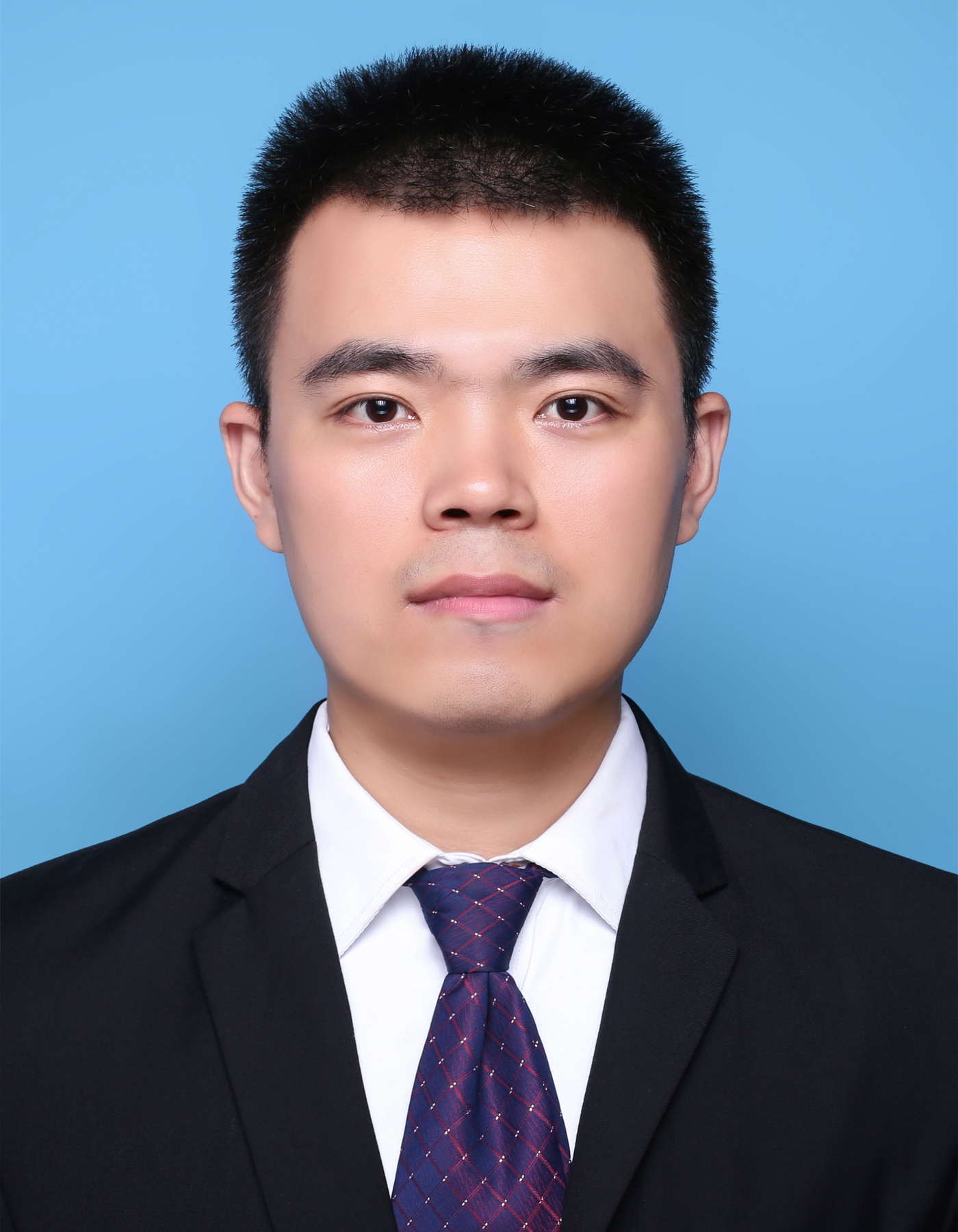}}]{Liwei Hu} received the B.S. degree in software engineering from the School of Software, Hebei Normal University, Shijiazhuang, Hebei, China, in 2014, the M.S. and Ph.D. degrees in computer science and technology from the University of Electronic Science and Technology (UESTC), Chengdu, China, in 2018 and 2023, respectively. He also
was a visitor at University of Copenhagen in 2023. He has been a lecturer in School of Information Science and Engineering, Hebei University of Science and Technology since 2024. His research fields include aerodynamic data modeling, geometric deep learning.
\end{IEEEbiography}

\begin{IEEEbiography}[{\includegraphics[width=1in,height=1.25in,clip,keepaspectratio]{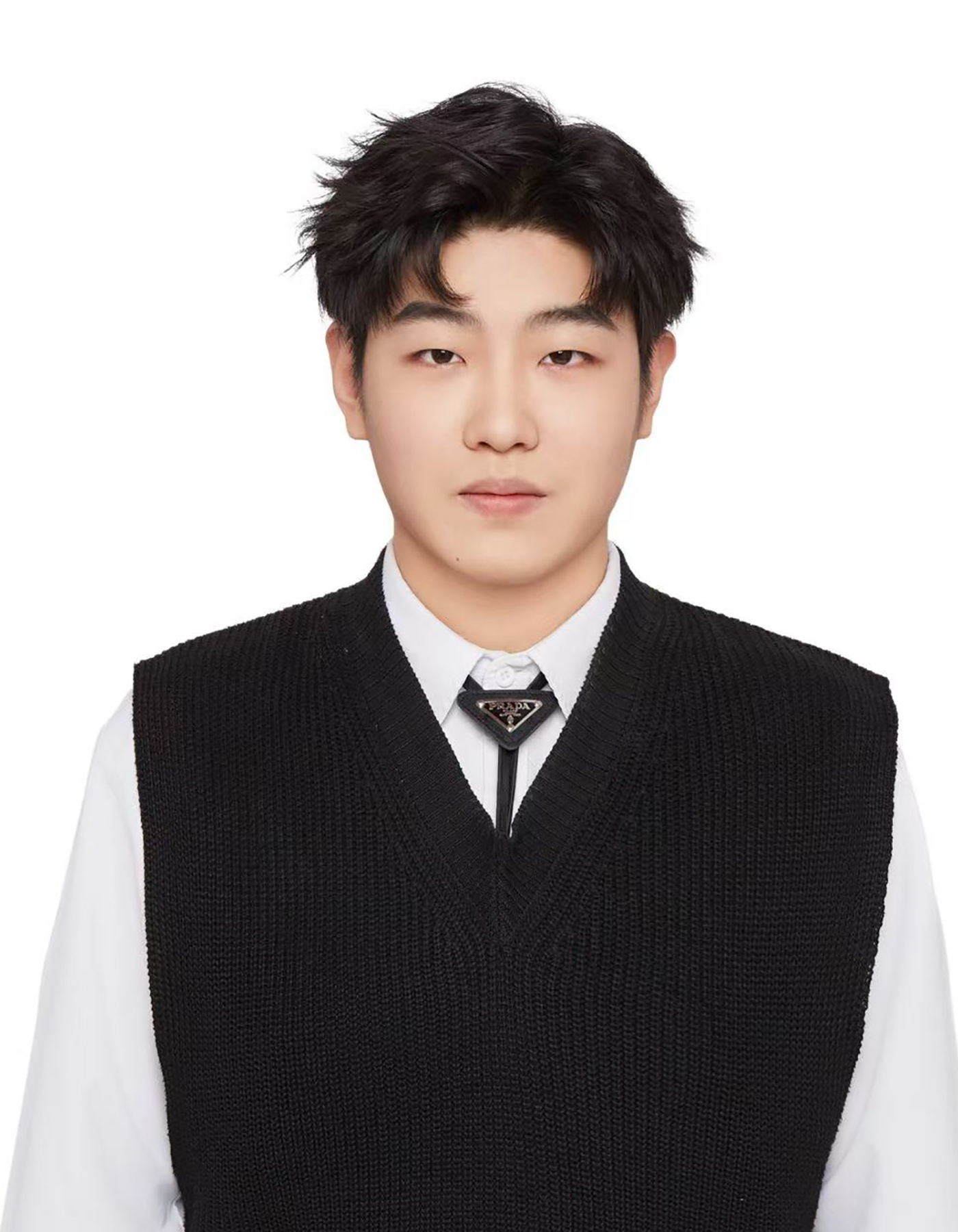}}]{Guangyao Li} received a B.S. degree in Computer Science and Technology from Hebei Agricultural University in 2024. He is pursuing a M.S. degree in Computer Science and Technology in Hebei University of Science and Technology. His research focus on the geometric deep learning. 
\end{IEEEbiography}

\begin{IEEEbiography}[{\includegraphics[width=1in,height=1.25in,clip,keepaspectratio]{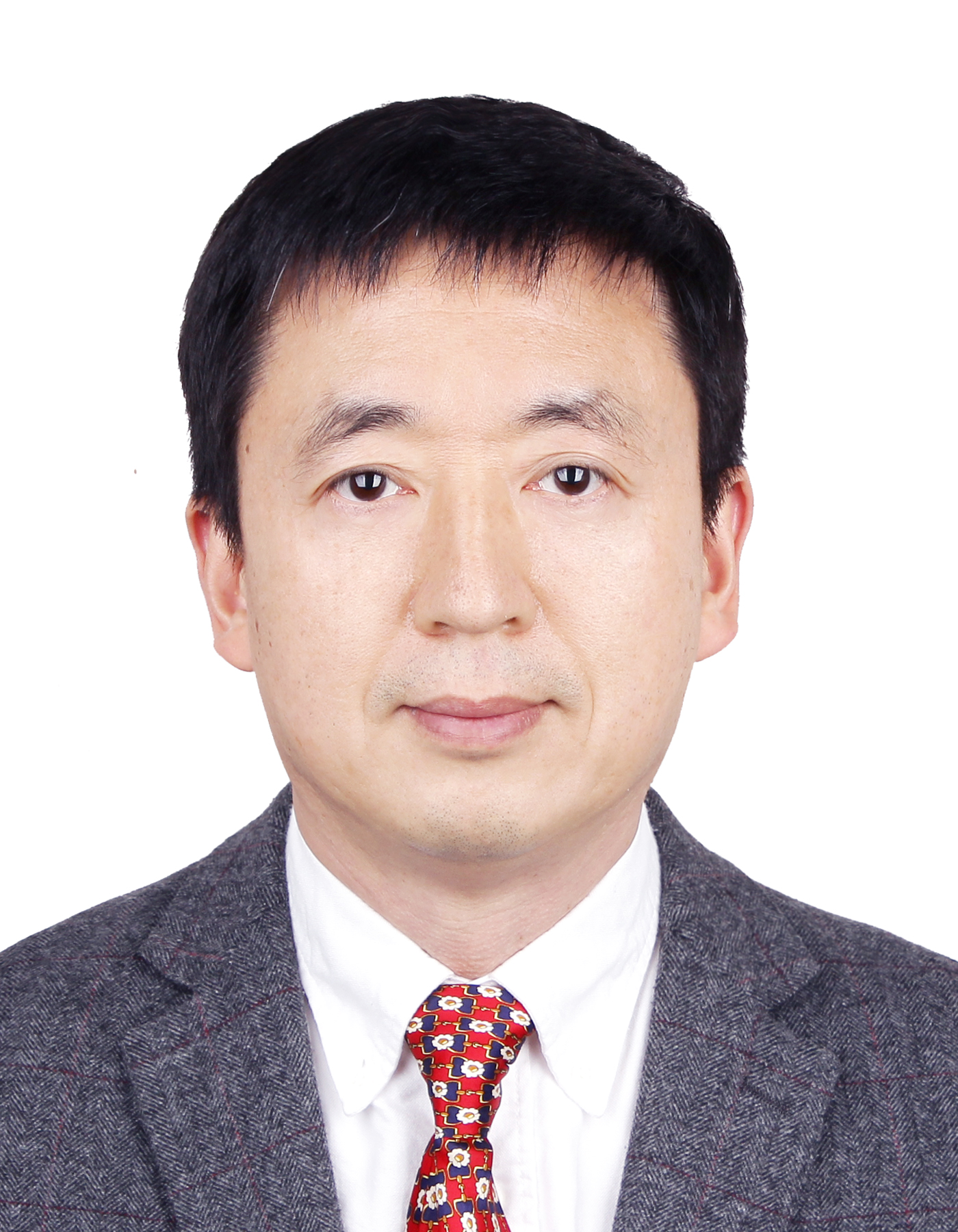}}]{Wenyong Wang} (Senior Member, IEEE) received the B.S. degree in computer science and technology from BeiHang University, Beijing, China, in 1988, and the M.S. and Ph.D. degrees in computer science and technology from the University of Electronic Science and Technology (UESTC), Chengdu, China, in 1991 and 2011, respectively. Since 2006, he has been a Professor in computer science and engineering at UESTC. He is also a special-term Professor with the Macau University of Science and Technology, Macau, China. His research interests include next generation Internet, software-defined networks, and software engineering. Dr. Wang is a Senior Member of the Chinese Computer Federation, and a Member of the expert board of the China Education and Research Network and China Next Generation Internet.
\end{IEEEbiography}

\begin{IEEEbiography}[{\includegraphics[width=1in,height=1.25in,clip,keepaspectratio]{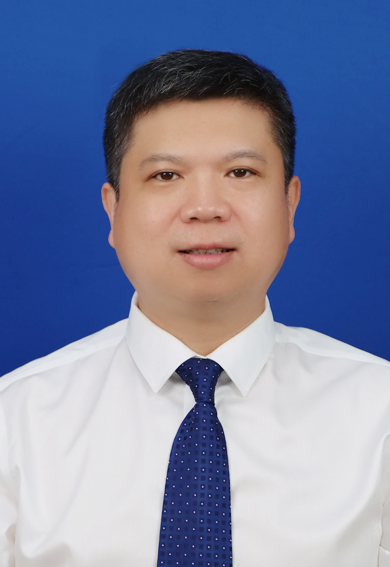}}]{Xiaoming Zhang} received the B.S. degree in computer science and technology from Hebei University of Science and Technology, Shijiazhuang, Hebei, China, in 1997, the M.S. in computer application technology from Hebei University, Baoding, Hebei, China, in 2002, and Ph.D. degree in computer application technology from the University of Science and Technology Beijing (USTB), Beijing, China, in 2009. Since 2016, he has been a Professor in computer science and technology  at Hebei University of Science and Technology. He is also a honorary Professor with the Federation University Australia, Ballarat VIC, Australia. His research interests include knowledge graph, knowledge-based systems and multi-modal knowledge fusion. Dr. Zhang is a Senior Member of the Chinese Computer Federation, and an Executive Committee Member of the Computer Application Special Committee of the China Computer Federation.
\end{IEEEbiography}

\begin{IEEEbiography}[{\includegraphics[width=1in,height=1.25in,clip,keepaspectratio]{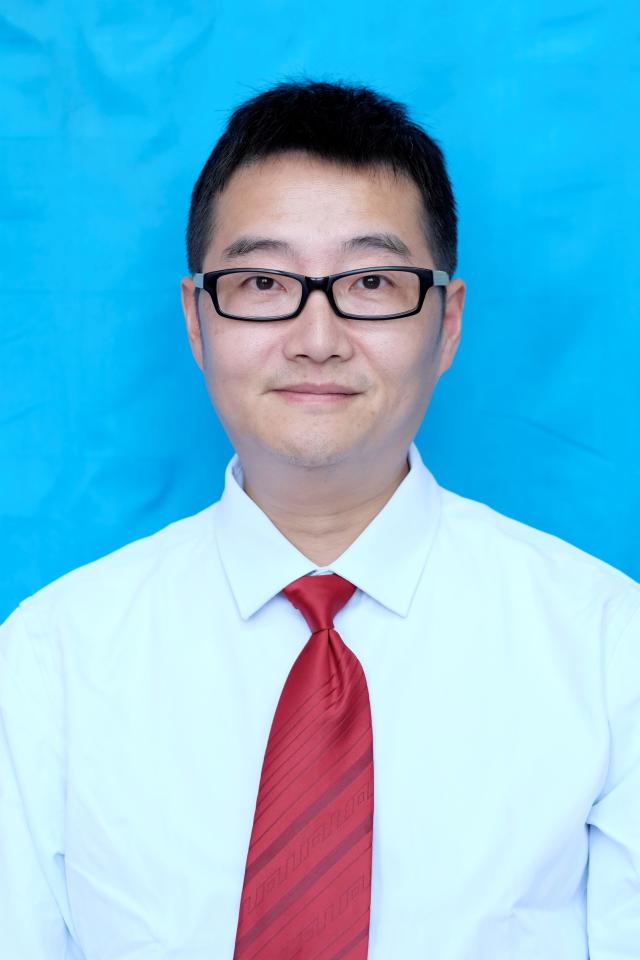}}]{Yu Xiang} (Member, IEEE) received the B.S., M.S., and Ph.D. degrees in computer science and technology from the University of Electronic Science and Technology of China (UESTC), Chengdu, China, in 1995, 1998, and 2003, respectively. He joined the UESTC in 2003 and became a Professor in 2024. From 2014 to 2015, he was a Visiting Scholar with the University of Melbourne, Parkville VIC, Australia. His current research interests include computer networks, intelligent transportation systems, and deep learning.
\end{IEEEbiography}

\end{document}